\documentclass[11pt, a4paper]{thuc3i}
\usepackage[compress]{natbib}
\setheadertext{One-Shot On-Policy Distillation}
\usepackage{booktabs}
\usepackage{tabularx}
\usepackage{multirow}
\usepackage[utf8]{inputenc} %
\usepackage[T1]{fontenc}    %
\usepackage{hyperref}       %
\usepackage{url}            %
\usepackage{tabularx}
\usepackage{makecell}
\usepackage{array,ragged2e,booktabs}
\newcolumntype{P}[1]{>{\RaggedRight\arraybackslash}p{#1}}
\usepackage{longtable}
\usepackage{amsfonts}       %
\usepackage{amsthm}         %

\usepackage{nicefrac}       %
\usepackage{microtype}      %
\usepackage{xcolor}         %

\usepackage{algorithm}
\usepackage{algorithmic}

\definecolor{darkblue}{rgb}{0, 0, 0.5}
\hypersetup{colorlinks=true, citecolor=darkblue, linkcolor=darkblue, urlcolor=darkblue}

\usepackage{xurl}

\usepackage{soul}
\usepackage{amsmath}
\usepackage{subcaption}
\usepackage{graphicx}
\usepackage{changes}
\usepackage{mathtools}
\usepackage{pifont}
\usepackage{tocloft}

\usepackage{colortbl}
\usepackage{wrapfig}
\usepackage{xspace}
\usepackage{multirow}
\usepackage{enumitem}

\usepackage{listings}
\usepackage{fontawesome5}

\usepackage{caption}

\usepackage{afterpage}
\usepackage{float}
\usepackage[edges]{forest}
\usepackage[normalem]{ulem}
\usepackage{bbding}
\usepackage[most,skins,theorems]{tcolorbox}
\usepackage{epigraph}

\usepackage[tikz]{bclogo}
\usepackage[framemethod=tikz]{mdframed}

\definecolor{lightorange}{HTML}{faa755}
\definecolor{lightblue}{RGB}{220,235,250}
\tcbset{
  takeawaysbox/.style={
    title=Takeaways,
    colback=lightblue!80,
    colframe=black,
    fonttitle=\bfseries\small,
    coltitle=white,
    colbacktitle=black,
    enhanced,
    attach boxed title to top left={xshift=2.5mm,yshift=-2.5mm},
    boxed title style={rounded corners, size=small, colframe=black, colback=black},
    width=\linewidth,
    arc=3.5mm
  },
  examplebox/.style={
    colback=orange!6,
    colframe=lightorange,
    fonttitle=\bfseries\small,
    coltitle=black,
    colbacktitle=orange!18,
    enhanced,
    attach boxed title to top left={xshift=2.5mm,yshift=-2.5mm},
    boxed title style={rounded corners, size=small, colframe=lightorange, colback=orange!18},
    width=\linewidth,
    arc=3.5mm
  }
}
\tcbset{
  taskbox/.style={
    title=Task Definition,
    colback=lightblue!80,
    colframe=black,
    fonttitle=\bfseries\small,
    coltitle=white,
    colbacktitle=black,
    enhanced,
    attach boxed title to top left={xshift=2.5mm,yshift=-2.5mm},
    boxed title style={rounded corners, size=small, colframe=black, colback=black},
    width=\linewidth,
    arc=2mm
  }
}

\definecolor{overviewbg}{HTML}{FFF3E0}
\definecolor{overviewframe}{HTML}{E65100}
\tcbset{
  overviewbox/.style={
    title=Overview,
    colback=overviewbg,
    colframe=overviewframe,
    fonttitle=\bfseries\small,
    coltitle=white,
    colbacktitle=overviewframe,
    enhanced,
    attach boxed title to top left={xshift=2.5mm,yshift=-2.5mm},
    boxed title style={rounded corners, size=small, colframe=overviewframe, colback=overviewframe},
    width=\linewidth,
    arc=3.5mm
  }
}

\definecolor{tplhead}{HTML}{1F2A6E}   % navy header bar
\definecolor{tplbodybg}{HTML}{EEEFF9} % light lavender (template-input tint)
\definecolor{tplkw}{HTML}{1F5FBF}     % code keyword blue
\definecolor{tplcmt}{HTML}{8A8A8A}    % code comment grey
\definecolor{tplnum}{HTML}{3A6EA5}    % line-number blue
\newtcolorbox{tplbox}[1]{%
  enhanced, colback=white, colframe=tplhead, colbacktitle=tplhead,
  coltitle=white, fonttitle=\bfseries\footnotesize, title={#1},
  boxrule=0.9pt, titlerule=0pt, arc=2pt,
  left=5pt, right=5pt, top=3pt, bottom=3pt,
  equal height group=tplex,   % both example boxes share max height (needs 2 compiles)
}
\newtcolorbox{tplinput}{%
  colback=tplbodybg, colframe=tplbodybg, boxrule=0pt, arc=1.5pt,
  left=4pt, right=4pt, top=2pt, bottom=2pt,
  before skip=3pt, after skip=3pt,   % tighten gap around the input box
}
\definecolor{tpltokctrl}{HTML}{1F5FBF}   % control token, e.g. <|begin_of_sentence|>
\definecolor{tpltokrole}{HTML}{9B2D86}   % role token, e.g. <|User|>
\definecolor{tpltokthink}{HTML}{1E8449}  % reasoning token, e.g. <think>
\definecolor{tpltokesc}{HTML}{8A8A8A}    % literal escape, e.g. \n
\newcommand{\tplctrl}[1]{{\color{tpltokctrl}\bfseries #1}}
\newcommand{\tplrole}[1]{{\color{tpltokrole}\bfseries #1}}
\newcommand{\tplthink}[1]{{\color{tpltokthink}\bfseries #1}}
\newcommand{\tplesc}[1]{{\color{tpltokesc}#1}}
\lstdefinestyle{tplpy}{%
  language=Python, basicstyle=\ttfamily\scriptsize,
  keywordstyle=\color{tplkw}\bfseries, commentstyle=\color{tplcmt},
  numbers=left, numberstyle=\tiny\color{tplnum}, numbersep=5pt,
  showstringspaces=false, breaklines=true, frame=none,
  aboveskip=2pt, belowskip=2pt, xleftmargin=12pt,
}

\definecolor{phenobg}{HTML}{EBF5FB}
\definecolor{phenoframe}{HTML}{2980B9}
\definecolor{mechbg}{HTML}{FEF5E7}
\definecolor{mechframe}{HTML}{E67E22}
\definecolor{recipebg}{HTML}{D5F5E3}
\definecolor{recipeframe}{HTML}{1E8449}
\definecolor{overviewdark}{HTML}{2C3E50}

\usepackage{cleveref}

\renewcommand{\today}{2026-09-04}

\usepackage[dvipsnames]{xcolor}

\title{Rethinking On-Policy Distillation of Large Language Models II: One Training Example}

\author{%
    Zixuan Fu$^{*1}$,
    Bingxiang He$^{*\dagger\ddagger1}$,
    Yuxin Zuo$^{*\dagger1}$,
    Haohuan Huang$^{*1,2}$,
    Jinqian Zhang$^{1}$,
    Ruhang Xiao$^{3}$,
    Cheng Qian$^{4}$,
    Qinyu Luo$^{5}$,
    Huan-ang Gao$^{1}$,
    Yudong Wang$^{1}$,
    Zhiyuan Liu$^{1}$,
    Ning Ding$^{\ddagger1}$,
    Chaojun Xiao$^{\ddagger1}$\\
    $^{1}$Tsinghua University \quad
    $^{2}$University of Chinese Academy of Sciences \quad
    $^{3}$Northeastern University\\
    $^{4}$University of Illinois Urbana-Champaign \quad
    $^{5}$Johns Hopkins University \quad
    \vskip -1mm
    \textbf{$^*$Equal Contribution.} \quad 
    \textbf{$^\dagger$Project Lead.} \quad \textbf{$^\ddagger$Corresponding Authors.} 
    \vskip -1mm
    \faGithub~{\fontfamily{ppl}\selectfont \textbf{Code:}}~\url{https://github.com/Thinking-Space/One-Shot-OPD}. \\
    \vskip 1mm
    \faEnvelope[regular] \texttt{hebx24@mails.tsinghua.edu.cn, \{dingning,xcj\}@tsinghua.edu.cn}
}

\begin{abstract}
    On-policy distillation (OPD) combines student-generated rollouts with dense token-level supervision from a teacher. Existing work has mainly studied its algorithmic behavior, leaving the role of training data unclear. We examine this role at the data-minimal limit by training on a single query. One-shot OPD keeps improving for hundreds of steps and recovers most of full-data OPD's gain across task domains and model families. We explain this result through the states visited during training and the rate at which the student aligns with the teacher. We measure \emph{state coverage}, the fraction of the states full-data OPD visits that a query set's rollouts reach. A single query already reaches \(71.5\%\), most of it within the first 100 steps. Adding semantically distinct queries raises coverage and validation accuracy together, until 16 queries reach \(98.9\%\) and match full-data training. Yet alignment slows at a similar pace whether OPD trains on one query or the whole dataset, and even a fixed set of states takes hundreds of steps to absorb. OPD is therefore data-overfed but algorithm-starved. Its rollouts quickly expose broad supervision, while the student absorbs that supervision increasingly slowly. The state-coverage result extends to multi-teacher OPD, where 16 semantically diverse queries per domain match full-data MOPD. As a further stress test, content-light templates and off-domain WildChat queries also approach the real-query baseline. Task content and induced state coverage can therefore come apart. We hope these findings direct future work toward the step efficiency of OPD, and prompt a re-examination of the data and the mechanisms behind its recent successes in frontier post-training.
\end{abstract}

\pretitlefigure{%
    \begin{center}
        \includegraphics[width=\linewidth]{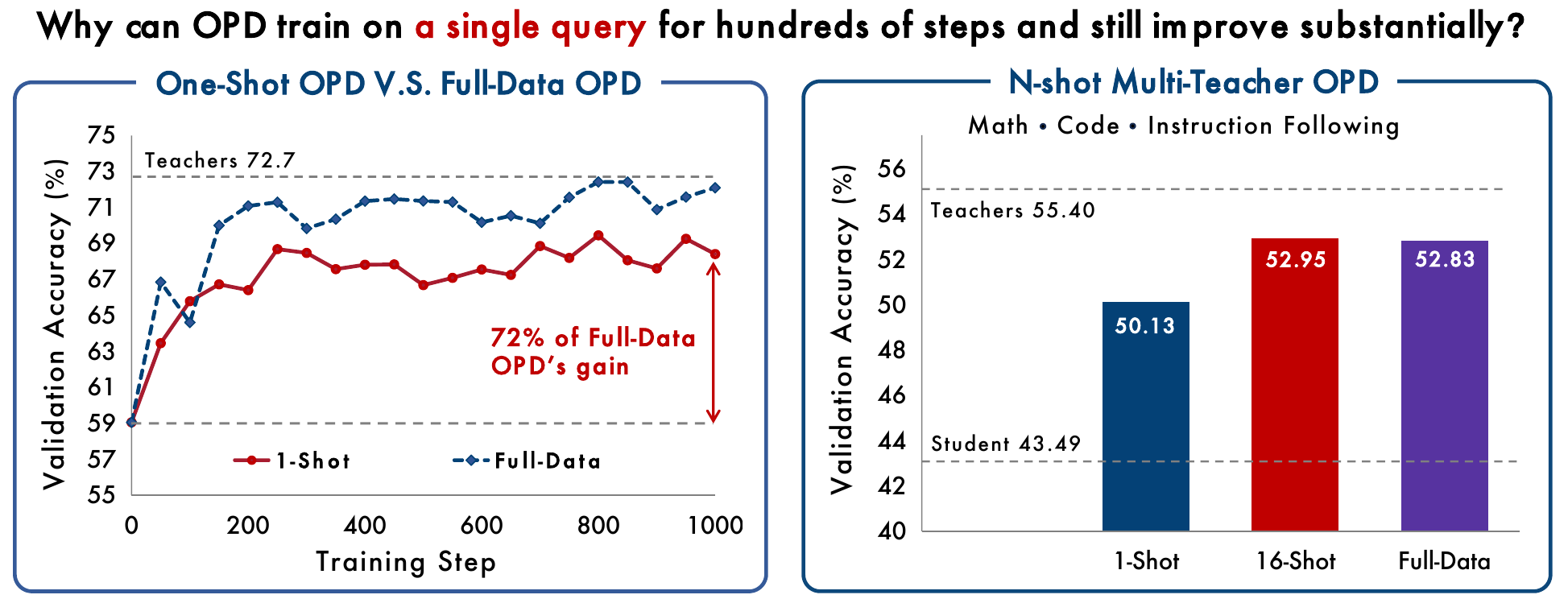}
        \captionof{figure}{%
            A single query recovers most of full-data OPD's gain on mathematical reasoning (left), and the effect holds across domains and model families. Under multi-teacher OPD, 16 queries per domain match full-data training (right).
        }
        \label{fig:teaser}
    \end{center}
    \vskip4pt
}

\begin{document}

\maketitle
\newpage

\section{Introduction}
\label{sec:intro}

On-policy distillation (OPD) is becoming increasingly common in frontier LLM post-training.
Qwen3~\citep{yang2025qwen3}, MiMo~\citep{xiao2026mimo}, GLM-5~\citep{zeng2026glm}, DeepSeek-V4~\citep{xu2026deepseek}, and Kimi K3~\citep{team2026kimi} all use OPD alongside supervised fine-tuning (SFT) and reinforcement learning (RL).
What makes OPD distinctive is its combination of on-policy state visitation and dense distillation supervision~\citep{gu2024minillm,agarwal2024policy}.
The student samples its own rollouts, while the teacher provides the full next-token distribution at every visited prefix, yielding dense, token-level supervision rather than the single outcome-level reward typically used in reinforcement learning with verifiable rewards (RLVR).

A growing line of work has systematically investigated OPD's training dynamics and mechanisms, largely from an algorithmic perspective, to explain how and why the method works~\citep{li2026rethinking,fu2026revisiting,cai2026learning,zhu2026many}.
Yet no prior work has studied how training data shapes OPD, or how the data and the algorithm interact.
Closing this gap is essential to a complete understanding of OPD.
In RLVR, \citet{wang2025oneshot} introduce one-shot RLVR as an extreme controlled experiment to isolate the role of training data.
We bring the same experimental lens to OPD by training on a single query, a setting we call one-shot OPD.
To our surprise, one-shot OPD produces a learning curve strikingly similar to that of one-shot RLVR as shown in Figure~\ref{fig:teaser}: despite repeatedly training on just one query, the model continues to improve for hundreds of steps and ultimately achieves a substantial gain.
We therefore devote this work to answering a single question:

\textit{Why can OPD, when trained on just one example, \textbf{keep learning for so long} and \textbf{improve by so much}?}

\begin{figure}[!ht]
    \centering
    \includegraphics[width=\linewidth]{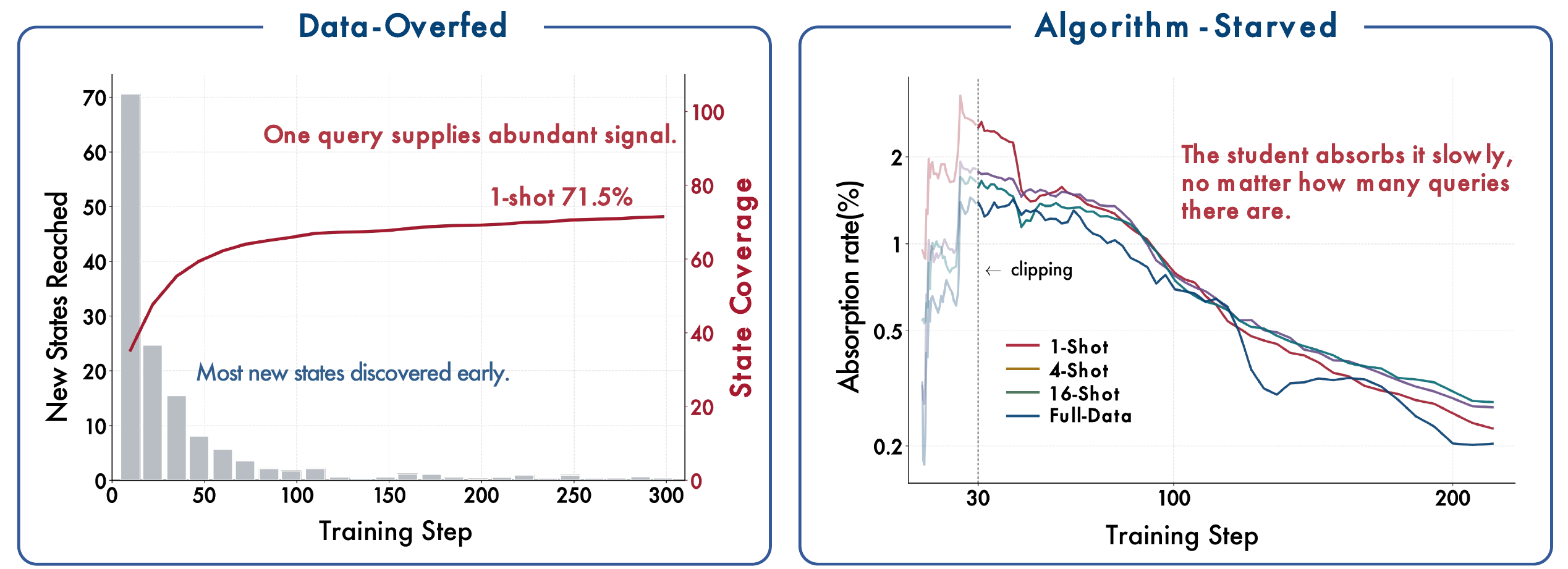}
    \caption{\textbf{OPD is data-overfed but algorithm-starved.} \textbf{Data:} one query already covers most of the states full-data OPD visits. \textbf{Algorithm:} the student absorbs an ever smaller proportion of the remaining teacher--student gap, no matter how many queries it trains on.}
    \label{fig:overview}
\end{figure}

This question exposes two aspects of OPD efficiency that standard training entangles:
\textit{\textbf{keep learning for so long}} concerns how fast the \textit{algorithm} absorbs supervision, whereas \textit{\textbf{improve by so much}} concerns why a small amount of \textit{data} supply sufficient supervision.
By reducing the data supply to its minimum, one-shot OPD disentangles the two, and the phenomenon itself proves robust: the gain holds across four task domains and three model families, and persists even on queries the student never solves (\Cref{sec:one-shot}).
Our answer is that \textbf{OPD is data-overfed but algorithm-starved}.
From the \textit{data perspective}, a query acts on the student through the \emph{states} its rollouts reach, each state being a prefix \(s=(x,y_{<i})\) at which the teacher supplies token-level supervision, so a query set is worth the part of that space it covers. We quantify this by \emph{state coverage}: we group the states full-data OPD visits into semantic clusters and report the fraction a setting's rollouts reach. A single query already supplies enough states on its own, covering \(71.5\%\) of what full-data OPD visits, and 16 semantically diverse queries cover \(98.9\%\) and match full-data training (\Cref{sec:data}). This is why so little data improves the student by so much. From the \textit{algorithm perspective}, what falls as training proceeds is the \emph{absorption rate}, the proportion of the remaining teacher--student gap that one update closes, and it declines in much the same way on one query as on all 17k (\Cref{sec:algorithm}), so the pace of a run is a property of OPD rather than of the training set.
In other words, the supervision a single query supplies remains largely unexploited, rate-limited by how quickly an on-policy student can absorb it.

The same mechanism informs data design beyond the one-shot setting.
In multi-teacher OPD (MOPD)~\citep{xiao2026mimo,xu2026deepseek}, where a single student is trained across several domains and each query is routed to its domain teacher, we find that 16 semantically diverse queries per domain suffice to match full-data training (\Cref{sec:mopd}).
Taking the data side to its extreme, we further show that even a training input that states no problem can remain effective: off-domain WildChat prompts and content-free templates drive OPD nearly as effectively as the full training set (\Cref{sec:template}).
Therefore, an input is useful largely because it starts the student reasoning.
Furthermore, we conduct a controlled comparison with one-shot RLVR on the same query to make the algorithmic contrast concrete: the outcome reward is exhausted once the query is solved on nearly every rollout, whereas OPD's token-level signal persists throughout training, and OPD's validation gain over $1000$ steps is more than twice that of RLVR (\Cref{sec:opd-vs-rlvr}).

In short, OPD is supplied with more supervision than its algorithm can absorb.
What data curation has to settle is therefore no longer how many problems to collect, but which states an input induces.
We hope this state-level view of the training data can guide future work on selecting queries by the states they induce, and on raising the rate at which a student absorbs them.

\section{Preliminaries}
\label{sec:prelim}

\subsection{Notation}
Let $x\sim\mathcal{D}$ denote a \emph{training input} and $y=(y_1,\ldots,y_L)$ a response, with $y_{<i}=(y_1,\ldots,y_{i-1})$ the prefix up to token position~$i$. We consider two LLMs: a student $\pi_{\theta}$ and a teacher $\pi_T$, each defining a next-token distribution over a shared vocabulary. A \emph{trajectory} is a pair $(x,y)$ with $y$ sampled autoregressively from a policy given $x$. We use $i$ to index token positions and $t$ to index optimization steps, writing $\theta_t$ for the student parameters at step~$t$. A \emph{state} is the autoregressive context $s_i=(x,y_{<i})$, the object on which the OPD objective is defined. 

\subsection{On-Policy Distillation}
On-policy distillation (OPD) samples trajectories from the student and aligns the student to the teacher on the prefixes the student actually visits~\citep{agarwal2024policy}. OPD can also be viewed as a special case of dense KL-constrained reinforcement learning, where the teacher distribution induces a token-level reward and the KL regularizer has a fixed relative weight~\citep{yang2026learning}.
 
In the full-distribution form, OPD minimizes a per-token KL divergence on visited states:
\begin{equation}
    \mathcal{L}_{\mathrm{OPD}}(\theta)
    =
    \mathbb{E}_{x\sim\mathcal{D},\,y\sim\pi_\theta}
    \left[
        \sum_{i=1}^{L}
        \operatorname{KL}\!\left(
            \pi_\theta(\cdot\mid s_i)
            \,\Vert\,
            \pi_T(\cdot\mid s_i)
        \right)
    \right].
\end{equation}

In practice, we estimate it with a per-token advantage:
\begin{equation}
    A_i^{\mathrm{OPD}} = \log \pi_T(y_i \mid s) - \log \pi_\theta(y_i \mid s),
    \label{eq:token-advantage}
\end{equation}

where $s = (x, y_{<i})$. Two properties follow from this formulation: (i)~the signal is \emph{local} in that it depends only on the prediction problem at state~$s$, not on the trajectory's outcome; and (ii)~it is \emph{dense} in that the teacher supplies a distributional signal at every visited state instead of a terminal scalar reward.

Equation~\ref{eq:token-advantage} evaluates the correction only at the token the student emitted. A second estimator of the same divergence keeps the student's $k$ most likely tokens at each visited state and weights them by the student's probability:
\begin{equation}
    A_i^{\text{top-}k}
    =
    \sum_{v \in \mathcal{V}_i}
    \tilde{\pi}_\theta(v \mid s)
    \left[
        \log \pi_T(v \mid s) - \log \pi_\theta(v \mid s)
    \right],
    \label{eq:topk-advantage}
\end{equation}
where $\mathcal{V}_i = \operatorname{TopK}(\pi_\theta(\cdot \mid s), k)$ and $\tilde{\pi}_\theta$ is $\pi_\theta$ renormalized over that set. It truncates the divergence to $k$ terms rather than estimating it from one sampled token, trading the distribution's tail for lower variance. Section~\ref{sec:setup} states which form each run uses. The alignment metrics below are diagnostics computed from the two policies' logged distributions, so they are available under either form.

\subsection{Dynamic Metrics}

\paragraph{Gap recovery.}
Because evaluation criteria and initial teacher--student performance gaps vary across domains and model pairs, raw score improvements are not directly comparable. We therefore report improvement as a ratio to the initial teacher--student gap. Let $M_0$, $M_T$, and $M_t$ denote the evaluation scores of the initial student, the teacher, and the student at optimization step~$t$, respectively. The gap recovery ratio at step~$t$ is
$\frac{M_t-M_0}{M_T-M_0}\times100\%.$
A value of $0\%$, $100\%$, or above $100\%$ indicates no improvement toward the teacher, matching the teacher, or surpassing the teacher, respectively.

\paragraph{Full-data recovery.}
When the reference is full-data OPD rather than the teacher, we normalize by the gain of full-data OPD instead. Let $M_F$ denote the full-data OPD score at the same optimization step. The full-data recovery ratio is
$\frac{M_t-M_0}{M_F-M_0}\times100\%.$ A value approaching $100\%$ means that a reduced query set nearly matches the gain of full-data OPD while a value above $100\%$ means it surpasses it.

\paragraph{Top-$k$ token overlap ratio.}
This metric measures agreement between the two policies' high-probability token sets. Let $S_t^\theta(s)=\operatorname{TopK}(\pi_{\theta_t}(\cdot\mid s),k)$ and $S^T(s)=\operatorname{TopK}(\pi_T(\cdot\mid s),k)$. The overlap at optimization step~$t$ is
\begin{equation}
    O_t^{(k)}
    =
    \frac{1}{|\mathcal{T}_t|}
    \sum_{i\in\mathcal{T}_t}
    \frac{|S_t^\theta(s_i)\cap S^T(s_i)|}{k},
    \label{eq:topk-overlap}
\end{equation}
where $\mathcal{T}_t$ indexes the non-padding response positions of the rollouts in the batch collected at step~$t$~\citep{li2026rethinking}.

\paragraph{Overlap-token advantage.}
Inside that agreement, we average the teacher--student log-probability difference over the shared tokens $S_t^\theta(s_i)\cap S^T(s_i)$, weighting each shared token by the probability the student assigns it. The result reports how far apart the two policies remain on the tokens they already rank highly. Because of that weighting, and because it averages over top-$k$ entries rather than over sampled tokens, it sits on a much smaller scale than the teacher--student distance of Section~\ref{sec:rate-setup}, and the levels of the two are not comparable.

\section{The One-Shot Phenomenon}
\label{sec:one-shot}

This section presents the experimental setup and results of One-Shot OPD. We investigate \textit{how much of full-data OPD's gain a single query recovers, and how robust that gain is}.

\begin{tcolorbox}[takeawaysbox]
\textbf{One query is enough to induce robust OPD gains.}
    A single query recovers most of the teacher--student gap across task domains and model families. The gain is insensitive to query difficulty, response length, and sampling temperature, and a query the student never solves is as effective as one it always solves.
\end{tcolorbox}

\subsection{Experimental Setup}
\label{sec:setup}

We establish one-shot OPD across math, code generation, instruction following, and agentic tool use for controlled analysis to test generality.

\paragraph{Models.}

\begin{table}[t]
\centering
\small
\caption{
Student--teacher pairs used in our experiments.
}
\label{tab:full}
\setlength{\tabcolsep}{5pt}
\renewcommand{\arraystretch}{1.12}
\begin{tabularx}{\textwidth}{@{}lXX@{}}
\toprule
\textbf{Setting} & \textbf{Student} & \textbf{Teacher} \\
\midrule

\multirow{3}{*}{Mathematical reasoning}
& DeepSeek-R1-Distill-Qwen-1.5B\newline
{\footnotesize (R1-Distill-1.5B)}
& JustRL-DeepSeek-1.5B\newline
{\footnotesize (JustRL-1.5B)} \\

& Llama-3.2-3B-Instruct\newline
{\footnotesize (Llama-3B-It)}
& GT-Llama-3.2-3B-Instruct-MATH\newline
{\footnotesize (GT-Llama-3B-Math)} \\

& OLMo-3-7B-Instruct-DPO\newline
{\footnotesize (OLMo-7B-It-DPO)}
& OLMo-3-7B-Instruct\newline
{\footnotesize (OLMo-7B-It)} \\

\midrule

Code generation
& DeepSeek-R1-Distill-Qwen-1.5B\newline
{\footnotesize (R1-Distill-1.5B)}
& Nemotron-Research-Reasoning-Qwen-1.5B-v2-RLVE\newline
{\footnotesize (Nemotron-1.5B)} \\

Instruction following
& DeepSeek-R1-Distill-Qwen-1.5B\newline
{\footnotesize (R1-Distill-1.5B)}
& UltraData-IF-1.5B \\

Agentic tool use
& Qwen2.5-Coder-1.5B-Instruct\newline
{\footnotesize (Qwen-Coder-1.5B)}
& Hammer2.1-1.5b\newline
{\footnotesize (Hammer-1.5B)} \\

\bottomrule
\end{tabularx}
\end{table}

For each domain, we pair a student with a post-trained teacher from the same family. Table~\ref{tab:full} lists the full identifiers and the short names used throughout. Math, code generation, and instruction following share the student R1-Distill-1.5B \citep{deepseekai2025deepseekr1incentivizingreasoningcapability}, whose teachers are, respectively, JustRL-1.5B \citep{he2025justrl}, Nemotron-1.5B \citep{zeng2025rlve}, and an instruction-following variant we post-train from the same student (Appendix~\ref{app:IF-setup}). For agentic tool use, the student is Qwen-Coder-1.5B \citep{hui2024qwen2} and the teacher is Hammer-1.5B \citep{lin2024hammer}. To test whether the phenomenon is specific to Qwen-based pairs, we add two mathematical-reasoning pairs from other families: Llama-3B-It \citep{grattafiori2024llama} with GT-Llama-3B-Math \citep{zhang2025co}, and OLMo-7B-It-DPO with OLMo-7B-It \citep{olmo2025olmo}.

\paragraph{Datasets.}
The domain-specific training sets comprise DAPO-Math-17K \citep{yu2026dapo} for mathematical reasoning, Open-R1 Codeforces \citep{penedo2025codeforces} for code generation, a sampled subset of UltraData-SFT-2605 \citep{ultradata-sft-2605} for instruction following, and xLAM-function-calling-60K \citep{liu2024apigen} for agentic tool use. For One-shot OPD in mathematics, we select three queries spanning easy, medium, and hard initial difficulty, scored by the student's pass rate over 8 rollouts before training (\(8/8\), \(4/8\), and \(0/8\), respectively). In each of the other domains, the one-shot query is sampled randomly from the corresponding training set. Appendix~\ref{app:model-data} details the dataset preprocessing and the selected one-shot queries.

\paragraph{Training.}
We implement OPD in \texttt{veRL} \citep{sheng2025hybridflow}.
The mathematical-reasoning runs of this section optimize the top-\(k\) advantage of Eq.~\ref{eq:topk-advantage} with \(k=16\); the code, instruction-following, and agentic runs optimize the sampled-token advantage of Eq.~\ref{eq:token-advantage}.
Every update uses a batch of $64$ rollouts. We use AdamW with a learning rate of $10^{-6}$, a rollout temperature of $1.0$, and a gradient clip norm of $1.0$; remaining hyperparameters are listed in Table~\ref{tab:opd-hyperparams}.

\paragraph{Evaluation.}
For math, we evaluate on MATH-500 \citep{hendrycksmath2021}, AMC 2023 \citep{li2024numinamath}, and AIME 2025 \citep{balunovic2026matharena}, sampling 16 responses per problem and reporting avg@16 accuracy.
For code generation, LiveCodeBench v6 (LCB v6) \citep{jain2025livecodebench} uses 3 sampled solutions per problem and reports avg@3 with the official execution-based evaluator.
For instruction following, Multi-IF \citep{he2024multi} evaluates three-turn conversations and reports the final-turn score averaged over its eight languages.
For agentic tool use, BFCL v3 \citep{patil2025bfcl} reports avg@8 over the evaluated subsets. And the response caps are \(31{,}744\) tokens
for mathematics, \(65{,}536\) for LCB v6, \(16{,}384\) per turn for Multi-IF, and \(4{,}096\) for BFCL v3. Unless stated otherwise, mathematics figures report validation accuracy macro-averaged over MATH-500, AMC 2023, and AIME 2025, and a dashed line marks the teacher.

\begin{figure}[t]
  \centering
  \includegraphics[width=\textwidth]{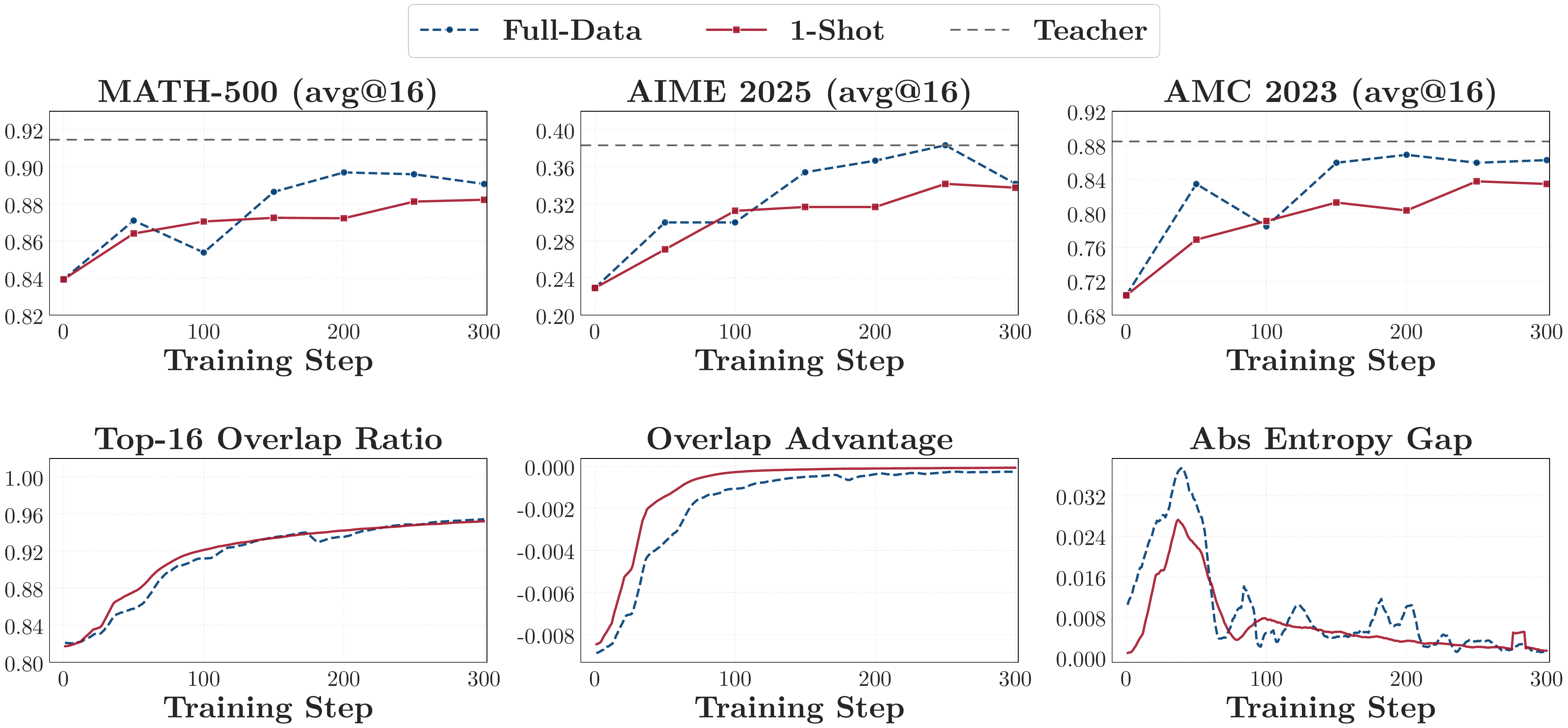}
  \caption{Validation accuracy and Training dynamics of one-shot OPD on mathematical reasoning.}
  \label{fig:math_oneshot_dynamics}
\end{figure}

\begin{figure}[t]
  \centering
  \includegraphics[width=\textwidth]{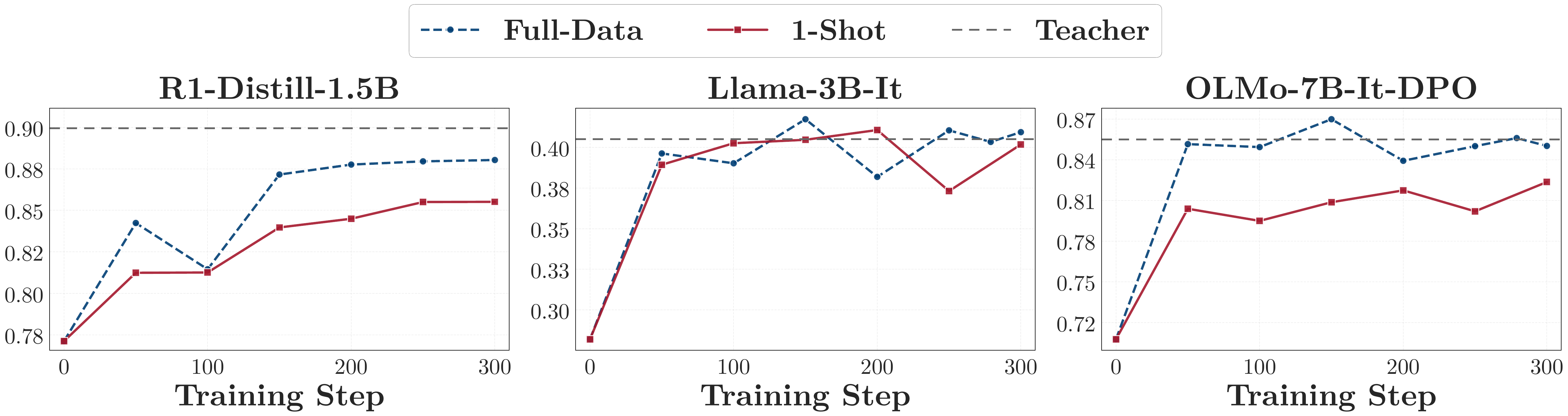}
  \caption{
  Validation accuracy of one-shot OPD across 3 student--teacher families, on mathematical reasoning averaged over MATH-500 and AMC 2023. AIME 2025 is excluded because the Llama pair scores near zero on it.
  }
  \label{fig:cross_family}
\end{figure}

\begin{figure}[t]
  \centering
  \includegraphics[width=\textwidth]{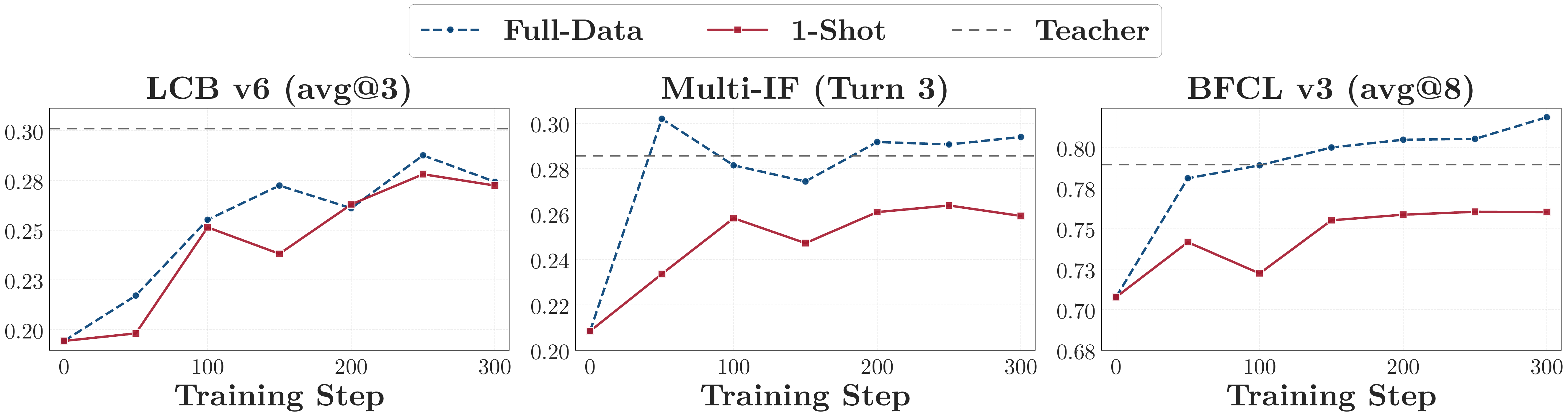}
  \caption{
  Validation accuracy of one-shot OPD across three task domains.
  }
  \label{fig:cross_domain}
\end{figure}

\begin{figure}[t]
  \centering
  \includegraphics[width=\textwidth]{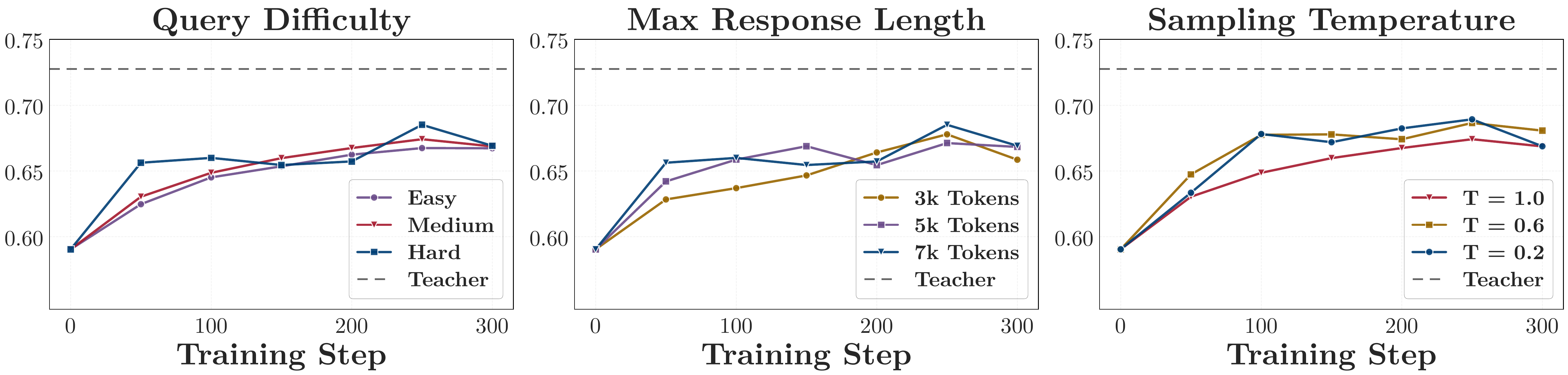}
  \caption{
  Validation Accuracy of one-shot OPD under varying query difficulty, response-length cap on the hard query, and sampling temperature on the medium query.
  }
  \label{fig:probe_prompt}
\end{figure}

\subsection{Experimental Results}
% \subsection{A Single Query Recovers Most of Full-Data OPD's Gain Robustly}
\label{sec:finding-recovery}

\paragraph{One-shot OPD recovers most of full-data OPD's gain in mathematics.}

Figure~\ref{fig:math_oneshot_dynamics} shows that training on a single query improves accuracy on MATH-500, AIME 2025, and AMC 2023, approaching the full-data OPD on all three benchmarks. Averaged over them, one-shot OPD reaches \(68.5\) against \(69.8\) for full-data OPD, recovering \(69\%\) of the teacher--student gap and \(87\%\) of full-data OPD's gain at step 300. Beyond step \(300\) both curves stay within a band of about \(3\) points, and the recovered fraction ranges from \(62\%\) to \(89\%\) through step \(1000\), where one-shot OPD reaches \(68.4\) against \(72.1\) and recovers \(72\%\) of full-data OPD's gain (Figure~\ref{fig:teaser}). The training dynamics follow the same alignment process under both settings: the top-16 overlap ratio climbs to the full-data level, the overlap-token advantage approaches zero, and the absolute entropy gap nearly closes. Thus, even when all rollouts originate from a single query, OPD improves accuracy while progressively aligning the student distribution with the teacher on the visited states.

\paragraph{The one-shot effect is robust across families and task domains.}
\label{sec:one-shot-robust}
Figure~\ref{fig:cross_family} shows that one-shot OPD improves mathematical reasoning across all three student--teacher families. At the final checkpoint, the averaged scores increase from \(77.1\), \(28.2\), and \(70.8\) for the respective student baselines to \(85.5\), \(40.2\), and \(82.4\) for R1-Distill-1.5B, Llama-3B-It, and OLMo-7B-It-DPO. Figure~\ref{fig:cross_domain} further shows that the effect extends beyond mathematical reasoning: on code generation, instruction following, and agentic tool use, one-shot OPD recovers \(73\%\), \(66\%\), and \(64\%\) of the corresponding teacher--student gaps, respectively. Together, these results show that the one-shot effect is robust across both model families and task domains.

\paragraph{The one-shot effect is robust to query and rollout properties.}

We next examine whether one-shot OPD remains effective under varying query difficulty, response-length budgets, and rollout sampling temperature. As shown in Figure~\ref{fig:probe_prompt}, one-shot OPD works across easy, medium, and hard queries, even though the fraction of correct rollouts evolves very differently in the three settings (Appendix~\ref{app:training-accuracy}): the easy query is solved on nearly every step, the medium query becomes largely solvable during training, and the hard query is never solved. Tightening the response-length cap and lowering the rollout temperature both preserve the one-shot gain. Together, these results show that one-shot OPD is robust to query difficulty, response-length budget, and rollout sampling temperature.

\section{Data Perspective: Abundant States}
\label{sec:data}

Section~\ref{sec:one-shot} showed that One-Shot OPD recovers much of full-data OPD's gain robustly. This section asks \textit{why training on a single query produces such a large gain, and whether the same explanation also covers the cases where n-shot OPD can match full-data OPD}. \footnote{The runs analyzed here and in Section~\ref{sec:algorithm} optimize the sampled-token advantage of Eq.~\ref{eq:token-advantage}, rather than the top-\(k\) advantage used by the mathematics runs of Section~\ref{sec:one-shot}, so their absolute accuracies differ slightly from those reported there. Every comparison below is between runs that share this form.}

\begin{tcolorbox}[takeawaysbox]
\begin{itemize}[topsep=0pt, partopsep=0pt, leftmargin=12pt, itemsep=0pt]
    \item \textbf{One query already reaches a large part of the full-data state space.}
    What OPD consumes is primarily states rather than queries. In practice, repeated rollouts from a single query can reach \(71.5\%\) state coverage, with most of it within the first \(100\) steps.

    \item \textbf{Diverse but few queries reach sufficient states and rival full-data performance.} Adding semantically distinct queries raises both state coverage and validation accuracy, and 16 queries already match full-data OPD. The value of an extra example lies in whether it covers states missed by earlier ones.
\end{itemize}
\end{tcolorbox}

\subsection{State Coverage of One-Shot OPD}
% \subsection{One Query Already Reaches \(71.5\%\) of the Full-Data State Space}
\label{sec:state-coverage}

\paragraph{Setup.}
OPD trains on states rather than queries. A query \(x\) and a sampled response \(y\) produce one state \(s=(x,y_{<i})\) at every token position \(i\), each paired with a target distribution from the teacher. With \(64\) rollouts per update, even a single query yields tens of thousands of supervised states, so query count can substantially understate the amount of supervision available to OPD.

\begin{tcolorbox}[examplebox, title={Hypothesis}]
\small
A single query can provide enough supervision to recover most of full-data OPD's gain if its rollouts cover a broad part of the state space visited by full-data OPD.
\end{tcolorbox}

To test this hypothesis, we measure the breadth rather than the raw number of visited states, since every generated token creates a distinct prefix. We call this measure \emph{state coverage}: we represent states in a shared representation space, partition that space into clusters, and report the fraction reached by a setting's rollouts.
\begin{itemize}[topsep=1pt, partopsep=0pt, leftmargin=13pt, itemsep=2pt, parsep=0pt]
    \item \textbf{Representation.} We represent each state by its \emph{teacher signature} \(h_T(s)\), the teacher's final-layer hidden vector at the state's last token.
    \item \textbf{Reference space.} We pool states visited by full-data OPD over its entire run on DAPO-Math-17K, sampling \(8\) uniformly spaced positions from each rollout. A held-out portion of these rollouts takes no part in defining the clusters and is measured as a setting of its own, \emph{full data (held-out)}.\footnote{Three of every five collected rollouts form the pool; the other two are the held-out setting, so it is what full-data OPD reaches on a fresh sample under the same budget.}
    \item \textbf{Clusters.} After PCA, \(K\)-means partitions the reference signatures into \(K=200\) clusters. Each state falls in the nearest one, denoted \(c(s)\).
\end{itemize}
The \emph{state coverage} of a set of states \(S\) is then the fraction of the \(K\) clusters it reaches:
\begin{equation}
    \mathrm{Cov}(S)
    =
    \frac{1}{K}
    \left|
    \left\{\,c(s)\;:\;s\in S\,\right\}
    \right|.
    \label{eq:state-coverage}
\end{equation}
Coverage records which clusters a setting reaches, not how often it visits them. We fit the clusters once and measure every setting over the same \(300\) steps. Full data (held-out) reaches \(100\%\) on this budget, so the top of the scale is a level full-data OPD attains rather than a maximum the construction guarantees. Appendix~\ref{app:state-coverage} gives the construction and schedule, and shows that the comparisons below are stable across choices of \(K\), reference set, and state positions.

\paragraph{One-shot OPD reaches \(71.5\%\) of the state space.}
\begin{figure}[t]
  \centering
  \includegraphics[width=0.75\linewidth]{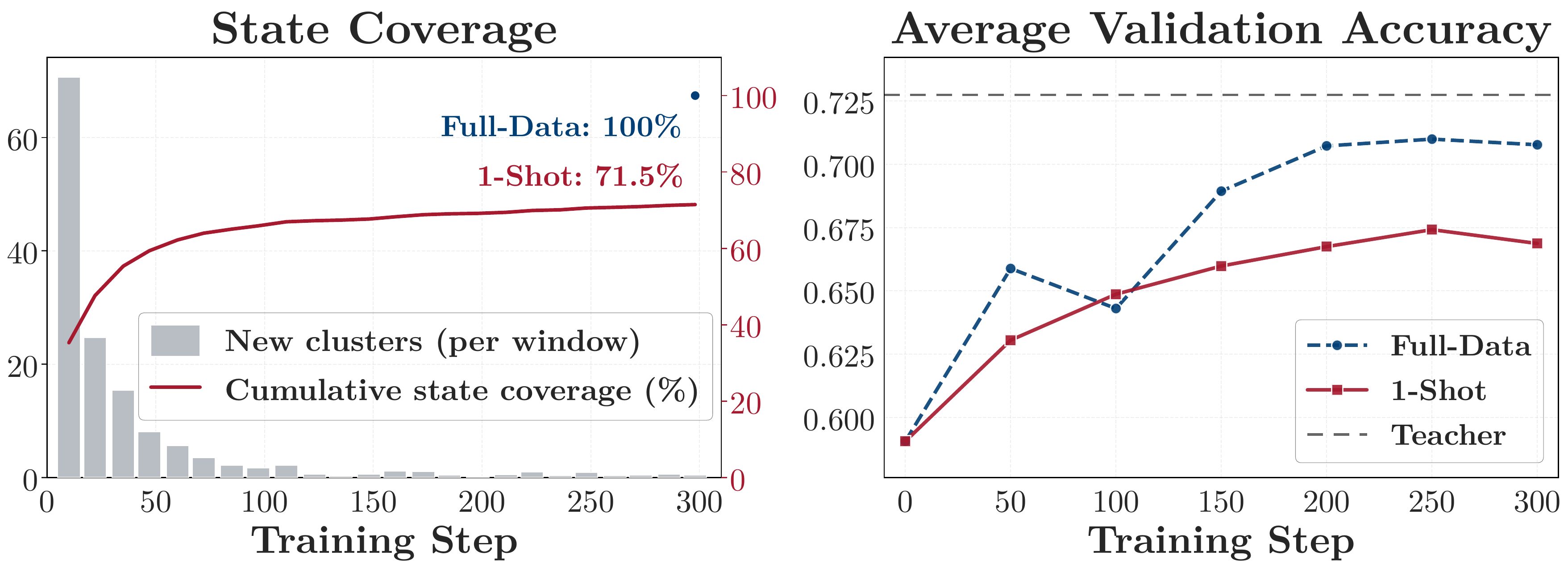}
  \caption{
    State coverage and validation accuracy of one-shot OPD on the medium query, against full-data OPD. Coverage is measured against the state space full-data OPD reaches, which is therefore the \(1.0\) level rather than a separate curve.
  }
  \label{fig:state-coverage}
\end{figure}
Figure~\ref{fig:state-coverage} shows that one-shot OPD reaches \(71.5\%\) state coverage by step \(300\). Most of this coverage appears early: the run reaches \(65.9\%\) by step \(100\), then adds only \(5.6\) percentage points over the next \(200\) steps. Repeated rollouts from the same query therefore continue to discover new clusters, but at a sharply diminishing rate.
The same run raises validation accuracy from \(59.1\) to \(66.9\), compared with \(70.8\) for full-data OPD at step \(300\). These results show the same pattern on the data side: a single query already covers most of the state-space clusters reached by full-data OPD and produces a substantial validation gain. One-shot OPD is therefore small in query count, yet broad in the supervision it generates.

\subsection{Diversity Expands State Coverage}
% \subsection{Diverse Data Can Reach More States and Train a Better Student}
\label{sec:state-diversity}

The analysis above shows that one query already reaches most of the full-data state space, but it leaves the causal question open: a run that trains well might simply visit more states along the way, with the extra states doing none of the work. We therefore ablate state coverage directly, varying how many distinct states the training data reaches while holding the rest of the setup fixed.

\paragraph{Setup.}
We raise data diversity from two directions, each with its own control.
\textbf{(i) Response diversity (off-policy).} We sample \(64\) trajectories once from the initial student on the one-shot query and keep that pool fixed, which also removes a confound of on-policy training, where the states keep changing as the student does. Nested subsets retain \(1\), \(4\), \(16\), or all \(64\) trajectories, and the retained ones are repeated to fill each batch of \(64\). Every condition shares the query, the batch size, and the optimization budget, so the number of distinct trajectories, and hence of states, is the only quantity that changes.
\textbf{(ii) Query diversity (on-policy).} These runs instead change the query set, and all follow the mathematical configuration. Starting from the medium query used by one-shot OPD, we build 4- and 16-shot sets by clustering DAPO-Math-17K with BGE-M3~\citep{chen2024bge} and taking one representative per semantic cluster, so each query added to the ladder is semantically distinct from those already in it, and we compare these sets with full-data OPD.
We report validation accuracy for every run, and state coverage for the on-policy ones, where the rollouts are those the student actually generates during training. Off-policy scores are averaged over the five evaluation checkpoints from step \(300\) to \(500\). On the coverage side, the full-data bar is the held-out split that sets the \(100\%\) ceiling.

\begin{figure}[t]
  \centering
  \includegraphics[width=\linewidth]{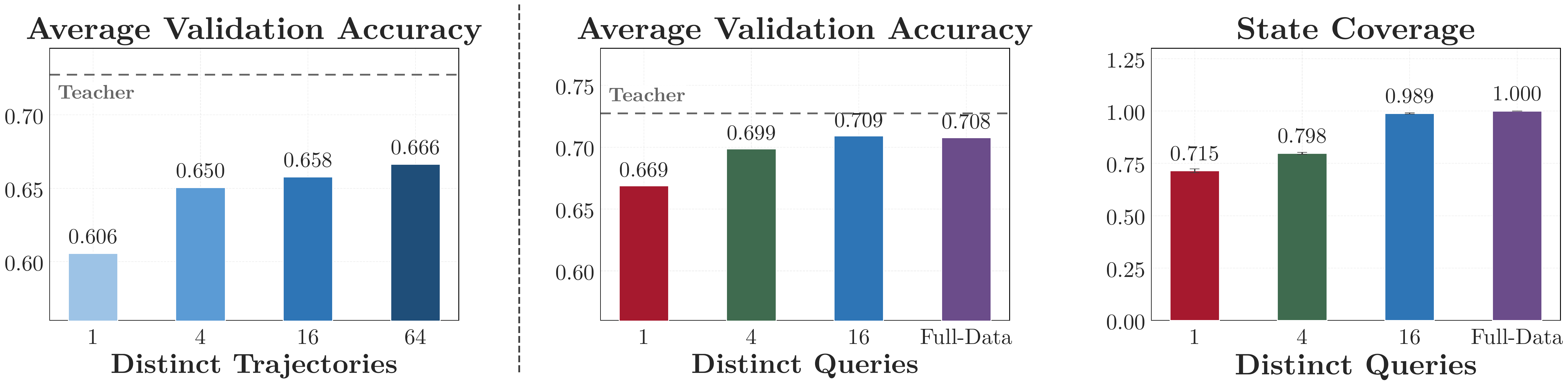}
    \caption{
    Validation accuracy and state coverage as the training data is made more diverse, on the response side by off-policy training on distinct trajectories and on the query side by on-policy training on distinct queries.}
  \label{fig:state-diversity}
\end{figure}

\paragraph{Data diversity is state diversity, and a few queries match full-data OPD.}
Figure~\ref{fig:state-diversity} shows off-policy validation accuracy rising monotonically with the number of distinct trajectories. Everything else is held fixed, so the only thing the extra trajectories add is states. The on-policy runs reach the same conclusion from the query side. Semantically distinct queries raise validation accuracy until 16 of them match full-data OPD, and they match it on state coverage at the same point, which rises from \(71.5\%\) for one query to \(98.9\%\) for 16. Every query on the ladder is a new semantic cluster, so it raises diversity and count together, and two controls in Appendix~\ref{app:fixed-count-diversity} separate them: holding the count at 16, drawing those queries from 16 semantic clusters rather than from one raises state coverage and validation accuracy sharply, whereas the order in which the student trains on a fixed query set does not matter. What an added query is worth is therefore set by whether it reaches new states, and a few diverse queries already reach nearly all the states full-data OPD does.
\section{Algorithm Perspective: Slow Alignment}
% The Absorption Rate Does Not Depend on the Training Set
\label{sec:algorithm}

Section~\ref{sec:data} analyzed the one-shot gain from the data side: what a query is worth to the student is set by the states its rollouts reach. Turning to the algorithm side, this section asks \textit{why one query keeps producing gains over so many steps}.

\begin{tcolorbox}[takeawaysbox]
\begin{itemize}[topsep=0pt, partopsep=0pt, leftmargin=12pt, itemsep=0pt]
    \item \textbf{Alignment slows over time, and this trend is insensitive to query count.} The student continuously narrows the teacher–student gap throughout the run, but the rate of progress diminishes steadily, which is why a run requires hundreds of steps rather than tens. Increasing the number of queries does not accelerate or decelerate this process.

    \item \textbf{Even a fixed set of states takes hundreds of steps to learn from.}
    A run that trains on the same states from beginning to end still improves over hundreds of steps, so a steady supply of fresh states is not required to keep a run going that long.
\end{itemize}
\end{tcolorbox}

\subsection{Alignment Metrics}
\label{sec:rate-setup}

OPD aligns the student with the teacher. Two quantities describe the state of that alignment at any moment: how much teacher--student disagreement is left, and how fast it is falling. We measure both on the states the student actually visits.

\begin{itemize}[topsep=2pt, partopsep=0pt, leftmargin=13pt, itemsep=2pt, parsep=0pt]
    \item \textbf{Distance.} The teacher and the student disagree at each visited position, by the per-token advantage of Eq.~\ref{eq:token-advantage}. We take the \emph{distance} \(d_t\) to be the average size of that disagreement over the positions \(\mathcal{T}_t\) the student visits at step \(t\):
    \begin{equation}
        d_t
        =
        \frac{1}{|\mathcal{T}_t|}
        \sum_{i\in\mathcal{T}_t}
        \left|
            \log\pi_T(y_i\mid s_i)-\log\pi_{\theta_t}(y_i\mid s_i)
        \right|.
        \label{eq:distance}
    \end{equation}
    Here \(s_i=(x,y_{<i})\) is the state at token position \(i\), as defined in Section~\ref{sec:prelim}. Averaging magnitudes keeps positive and negative per-token gaps from cancelling, so \(d_t\) falls to zero only as the two policies come to agree at the positions the student visits.
    \item \textbf{Absorption rate.} The \emph{absorption rate} \(v_t\) is then the proportion of that distance one update absorbs,
    \begin{equation}
        v_t = \frac{d_t-d_{t+1}}{d_t}.
        \label{eq:absorption-rate}
    \end{equation}
\end{itemize}
We track both metrics from step 30, after gradient clipping no longer binds. Since each run measures distance over the states it visits, these runs have settled at different levels by then. We therefore report each run against its own value at step 30, which we refer to as the \emph{distance left} \(d_t/d_{30}\).\footnote{Late in training a single pair of consecutive steps is too noisy to read, so \(v_t\) is estimated over the geometric window \([t/1.4,\,1.4t]\), which is why it stops short of step \(300\). Appendix~\ref{app:alignment-dynamics} gives the estimators and the evidence behind both conventions.}

\subsection{Alignment Slows Similarly Across Data}
\label{sec:mech-dynamics}

We track \(d_t\) and \(v_t\) for OPD trained on 1, 4, 16, and all DAPO-Math-17k queries, over the same \(300\) steps and under the mathematical configuration of Section~\ref{sec:setup}.

\begin{figure}[t]
    \centering
    \includegraphics[width=0.75\linewidth]{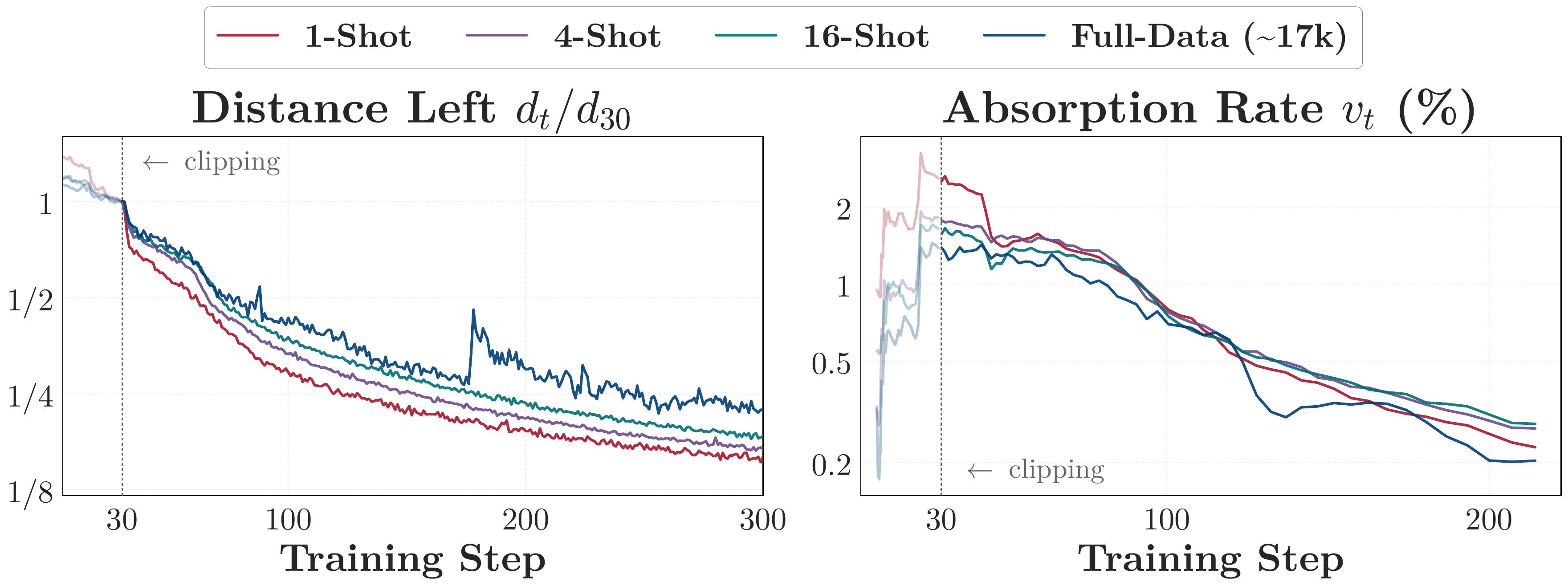}
    \caption{
    Teacher--student alignment for OPD trained on 1, 4, 16, and all DAPO-Math-17k queries. Distance (left) and absorption rate (right) are both on logarithmic vertical axes against linear steps.
    }
    \label{fig:alignment-contraction}
\end{figure}

\paragraph{The student aligns ever more slowly.}
The distance falls for the whole run, so the student is never stuck. What slows is the absorption rate, which falls throughout the run in the right panel of Figure~\ref{fig:alignment-contraction}: on the logarithmic distance axis at left a constant rate would trace a straight line, and all four curves bend flatter instead. Each update thus absorbs less of what is left than the last, which is why a run takes hundreds of steps rather than tens.

\paragraph{The absorption rate falls in much the same way at every training-set size.}
The four runs do this alike: each removes \(78\%\) to \(84\%\) of its step-\(30\) distance by step \(300\), and all four slow by a similar factor during training (Appendix~\ref{app:alignment-dynamics} gives the per-run numbers and the conventions these readings are taken under). One query therefore does not make alignment slower or faster: the pace is a property of OPD rather than of the training set. The learning rate changes how many steps that takes, but not the way the distance falls (Appendix~\ref{app:lr-sensitivity}).

\subsection{Ablation on Fixed States}
\label{sec:fixed-states}
On-policy training keeps producing new states as the student changes, and one reading of the hundreds of steps is that the run lasts as long as that supply does. Holding the training states fixed tests that reading directly: if a fixed set of states is exhausted quickly, the off-policy run should finish early.

\paragraph{Setup.}
We compare two one-shot runs that differ only in where their training states come from. The always-on-policy run draws fresh rollouts from the current student at every step. The always-off-policy run reuses the \(64\) trajectories of Section~\ref{sec:state-diversity}, sampled once from the initial student, for every update, so its training states stay fixed while the student changes.

\paragraph{Holding the states fixed leaves the run just as long.}

\begin{figure}[t]
    \centering
    \includegraphics[width=\linewidth]{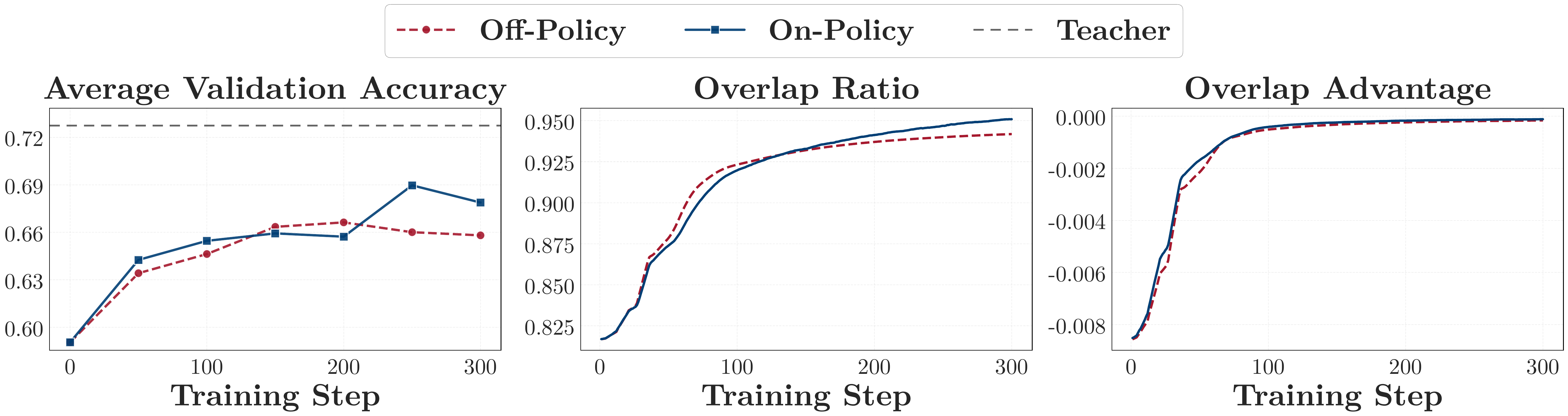}
    \caption{
    Validation accuracy and alignment metrics for the on-policy and off-policy one-shot runs.
    }
    \label{fig:online-offline}
\end{figure}
Figure~\ref{fig:online-offline} shows the off-policy run gains accuracy steadily for about \(200\) steps and then stops, so its gain builds up over hundreds of steps rather than over the first few updates. The alignment metrics move just as slowly: the off-policy run's top-\(16\) token overlap rises and its overlap-token advantage approaches zero only over hundreds of steps. Removing that supply therefore does not shorten the run. 
% The off-policy run is even slightly ahead at steps \(150\) and \(200\), and ends \(2.1\) points below on-policy training at step \(300\), across three benchmarks. Fresh states therefore buy a modest gain in final accuracy rather than a faster run. 
This suggests that a fixed set of states is enough on its own to keep an OPD run going for hundreds of steps, so the length of a run is not explained by how long fresh states keep arriving.

\section{One-Shot Extends to Multi-Teacher OPD}
% 16 Diverse Queries Remain Sufficient in Multi-Teacher OPD
\label{sec:mopd}

Section~\ref{sec:state-diversity} showed that a small, semantically diverse query set can match full-data OPD within one domain. Modern post-training pipelines often use \emph{multi-teacher on-policy distillation} (MOPD)~\citep{xiao2026mimo,xu2026deepseek}, where one student is trained on several domains in a single run and each query is routed to its domain teacher. This section asks \textit{whether the same query-diversity result carries over: can a small, diverse query set for each domain match full-data MOPD?}

\begin{tcolorbox}[takeawaysbox]
% \begin{itemize}[topsep=0pt, partopsep=0pt, leftmargin=12pt, itemsep=0pt]
%     \item 
    16 semantically diverse queries within each domain are sufficient to match full-data MOPD, which reaches nearly the same average validation accuracy as separate full-data OPD.
% \end{itemize}
\end{tcolorbox}

\begin{figure}[t]
    \centering
    \includegraphics[width=\textwidth]{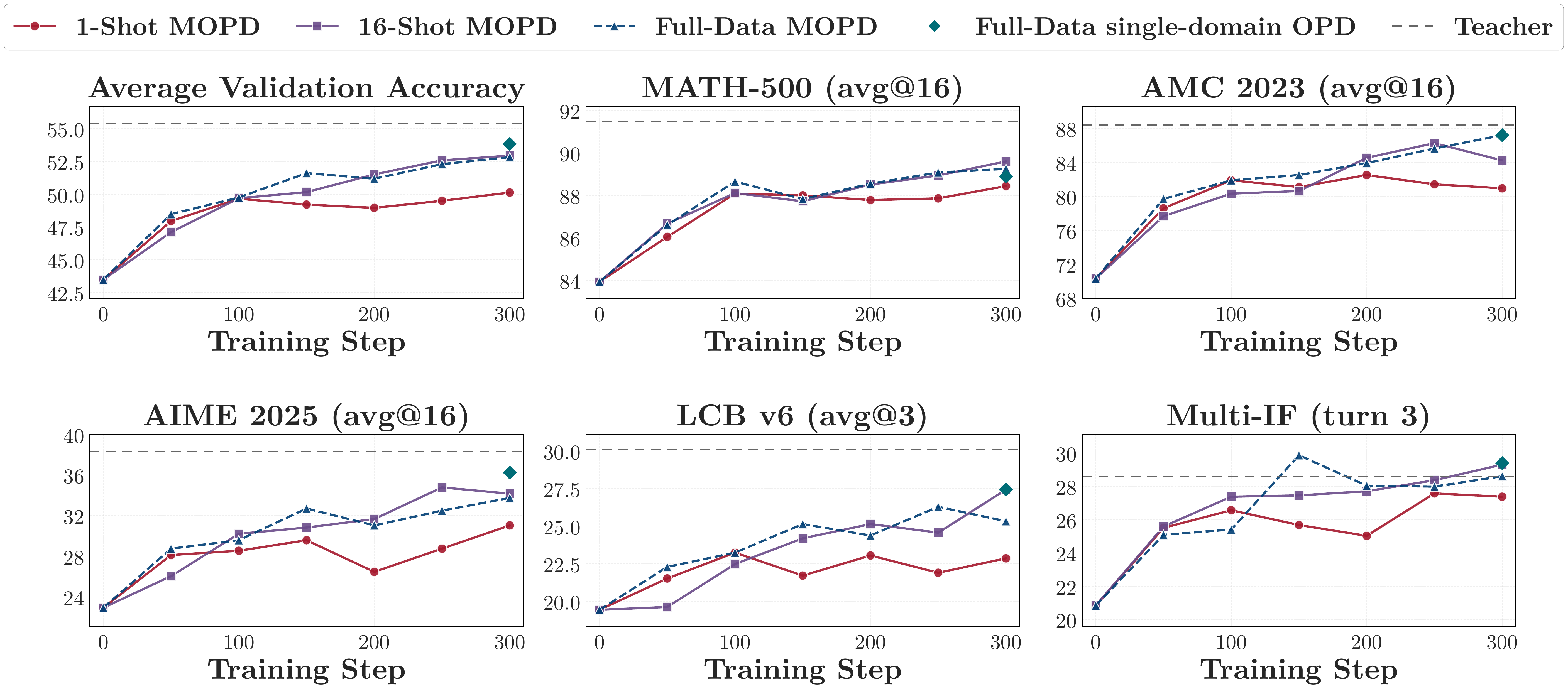}
    \caption{
    Validation accuracy of one-shot, 16-shot, and full-data MOPD.
    }
    \label{fig:mopd}
\end{figure}

\paragraph{Setup.}
We train one R1-Distill-1.5B student with MOPD on mathematical reasoning, code generation, and instruction following. The domain teachers from Section~\ref{sec:setup} are JustRL-1.5B for mathematics, Nemotron-1.5B for code, and UltraData-IF-1.5B for instruction following. We exclude agentic tool use because it uses a different student, Qwen-Coder-1.5B. The three settings are one-shot, 16-shot, and full-data MOPD. As in Section~\ref{sec:state-diversity}, the 16-shot setting selects one representative from each of 16 BGE-M3~\citep{chen2024bge} semantic clusters fitted separately within each domain. All settings run for \(300\) steps with the hyperparameters in Table~\ref{tab:mopd-hyperparams}.

\paragraph{Full-data MOPD matches separate full-data OPD.}
At step \(300\), full-data MOPD raises the average validation accuracy from \(43.5\) to \(52.8\), recovering \(79\%\) of the teacher--student gap. The three corresponding full-data OPD runs reach \(53.8\). This \(1.0\)-point difference shows that MOPD remains effective when all three domains are trained in a single run and establishes full-data MOPD as the reference for the query-diversity comparison below.

\paragraph{16 diverse queries per domain match full-data MOPD.}
Increasing the query count from 1 to 16 per domain raises the average validation accuracy from \(50.1\) to \(52.9\) at step \(300\). Full-data MOPD reaches \(52.8\), so 16-shot MOPD recovers \(101\%\) of full-data MOPD's gain. The result is not driven by one domain: calculated separately, the corresponding proportions are \(93\%\) for mathematics, \(136\%\) for code generation, and \(109\%\) for instruction following. Because 16-shot and full-data MOPD use the same rollout and optimization budgets, expanding each domain's query set beyond 16 diverse queries produces no additional average gain in this setting.

Together, these results are consistent with a state-space account for MOPD: semantically diverse queries can induce complementary states within each domain, allowing 16 queries per domain to match full-data MOPD.

\section{Discussion}
\label{sec:discussion}

The mechanism above shifts attention from prompts to the regions of reasoning states they induce. An input is useful when the student's rollouts reach regions where the teacher can provide useful supervision. Stating a problem is only one way to reach them. This view raises two broader questions: how much task content must the input itself contain, and why does the same one-shot setting behave differently under OPD and RLVR? We discuss these questions through content-light and off-domain inputs, followed by a controlled comparison with one-shot RLVR.

\subsection{Inputs as State Generators}
\label{sec:template}

The state-generator view suggests that task content may be unnecessary if an input still induces trajectories that the teacher can supervise. We stress-test this prediction with an empty user turn ending in \texttt{<think>}, the same template with a simple domain system prompt, and general-domain queries from WildChat~\citep{zhao2024wildchat} (Figure~\ref{fig:template-examples}). The WildChat set contains \(192{,}824\) predominantly English queries from general conversation, creative writing, role-playing, and rewriting. Conservative heuristics label only \(0.17\%\) as mathematics-related and \(2.63\%\) as code-related.

\begin{figure}[t]
\centering
\begin{minipage}[t]{0.485\textwidth}
\begin{tplbox}{(a) Base template}
\begin{tplinput}
\ttfamily\scriptsize\raggedright \tplctrl{<|begin\_of\_sentence|>}\tplrole{<|User|>}\tplthink{<think>}\tplesc{\textbackslash n}
\end{tplinput}
\end{tplbox}
\end{minipage}\hfill
\begin{minipage}[t]{0.485\textwidth}
\begin{tplbox}{(b) System template --- Code}
\begin{tplinput}
\ttfamily\scriptsize\raggedright \tplctrl{<|begin\_of\_sentence|>}Solve competitive programming problems. Read from stdin, write to stdout.\tplrole{<|User|>}\tplthink{<think>}\tplesc{\textbackslash n}
\end{tplinput}
\end{tplbox}
\end{minipage}
\\[4pt]
% {\scriptsize\tplctrl{\ttfamily control token}\quad\tplrole{\ttfamily role token}\quad\tplthink{\ttfamily reasoning token}\quad\tplesc{\ttfamily literal newline}\quad{\ttfamily instruction text}}
\caption{The two content-light templates used as OPD training inputs.}
\label{fig:template-examples}
\end{figure}

\begin{figure}[t]
\centering
\includegraphics[width=\textwidth]{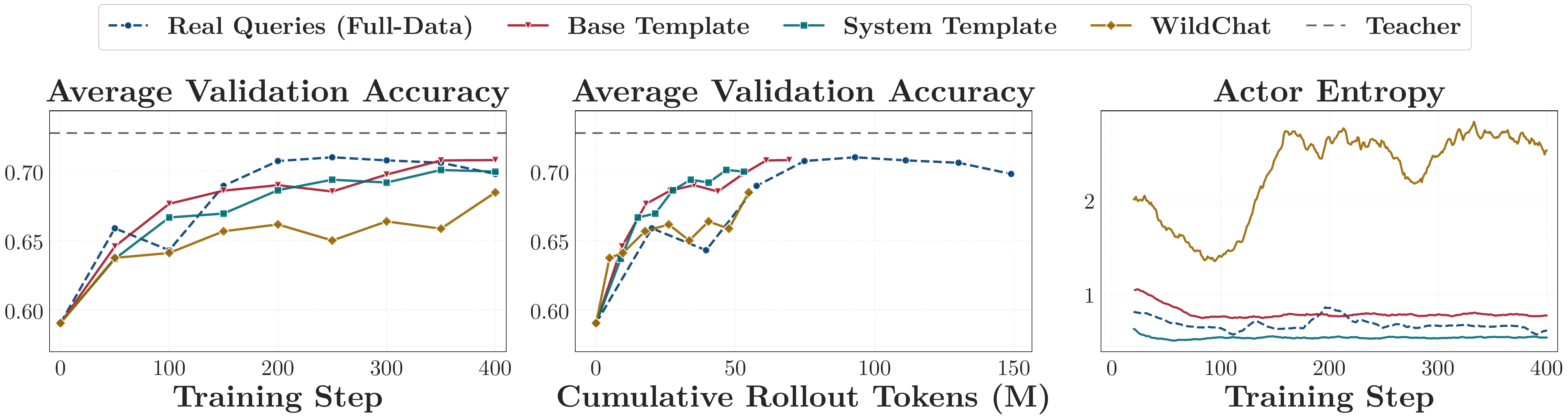}
\caption{
Validation accuracy and actor entropy for mathematics OPD trained on the real training set, on a content-light template, or on general-domain WildChat queries.
}
\label{fig:template-wildchat}
\end{figure}

\paragraph{Content-light and off-domain inputs can still induce useful states.}
% \paragraph{The input does not have to state a problem, even domain.}
All three conditions track the real-query baseline, which lifts the three-benchmark average from \(59.1\) to \(69.8\), and all three reach that level on a third to a half of its rollout tokens (Figure~\ref{fig:template-wildchat}). Code generation gives the same picture (Appendix~\ref{app:template-code}). 
However, the requirement is not empty: a scaffold that closes the block at once, \texttt{<think>\textbackslash n}\allowbreak\texttt{</think>\textbackslash n}, collapses into short meta-level replies we cannot train on.
These results do not imply that task content is generally dispensable. They show that content is not the only source of useful signal in OPD. An input also acts as a state generator, and even content-light or off-domain inputs can sometimes lead the student to states that support useful teacher supervision. Explicit domain content changes final performance by about one point in these experiments. This provides a complementary view of OPD data efficiency. Data quality depends not only on the problems being collected, but also on the state regions they induce and the teacher targets available there.

\subsection{Relation to One-Shot RLVR}
\label{sec:opd-vs-rlvr}

\begin{figure}[t]
\centering
\includegraphics[width=\textwidth]{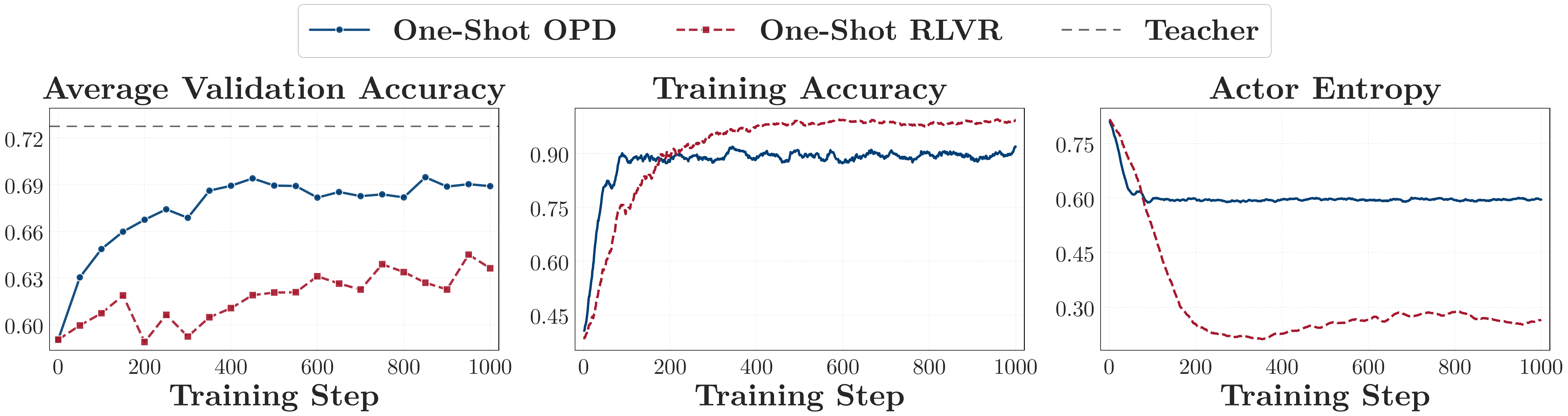}
\caption{
Validation accuracy and training dynamics for one-shot OPD and one-shot RLVR on the same query. Both take a batch of \(64\) queries per step. RLVR draws \(8\) rollouts per query and scores each by outcome under GRPO. OPD draws \(1\) and scores every sampled token against the teacher. Training accuracy is the fraction of rollouts that solve the training query.
% which RLVR optimizes and OPD only measures.
}
\label{fig:opd-vs-rlvr}
\end{figure}

One-shot OPD and one-shot RLVR~\citep{wang2025oneshot} share the same surface phenomenon that a single query supports continued improvement over hundreds of updates in OPD and over a thousand updates in RLVR. The source of learning is different. OPD receives a teacher target at every visited state, while RLVR receives a verifiable outcome for each trajectory. We compare them on the same medium query to isolate this difference (Figure~\ref{fig:opd-vs-rlvr}).

\paragraph{Useful inputs differ across objectives.}
RLVR updates from trajectory outcomes. Its training input must therefore define a task whose sampled outcomes can be verified. The outcomes must also vary enough to produce a useful advantage. OPD instead updates from the teacher--student gap at each visited state. It can receive useful signal whenever a prompt leads to trajectories that expose such gaps. The value of an OPD input is therefore not tied directly to verifiability or difficulty. Our results further show that even weak domain alignment can work in the settings we study. Coverage of the induced state regions still matters (Section~\ref{sec:data}), but obtaining such coverage is less restrictive than obtaining tasks with verifiable outcomes.

\paragraph{OPD extracts more signal from the same query.}
Both algorithms obtain fresh rollouts from the training query at every step. Over \(1000\) steps, OPD closes \(72\%\) of its gap to the teacher. Its gain in validation accuracy is more than twice that of RLVR. This advantage remains when the two methods are matched by rollout tokens rather than training steps (Figure~\ref{fig:opd-vs-rlvr}).

The RLVR signal weakens as the model learns to solve the training query. Its groups soon become nearly unanimous, which leaves no advantage under GRPO. OPD does not depend on variation in trajectory outcomes. It can continue learning from local teacher--student gaps after the query is solved, although these gaps gradually shrink. One-shot RLVR is therefore limited by the outcome variation available from its query. One-shot OPD is instead limited by how quickly it absorbs dense teacher supervision, as shown in Sections~\ref{sec:data} and~\ref{sec:algorithm}. The objectives also imply different ceilings. RLVR is not tied to a teacher's distribution, while OPD is trained to match one.

\section{Related Work}
\label{sec:related}

\paragraph{On-policy distillation.}
On-policy distillation (OPD) trains the student on its own rollouts with teacher supervision~\citep{song2026survey}. MiniLLM~\citep{gu2024minillm} optimized this under reverse KL by policy gradient, and GKD~\citep{agarwal2024policy} generalized it across divergences and on- and off-policy mixtures. \citet{yang2026learning} cast OPD as dense KL-constrained RL whose per-token log-ratio acts as an implicit reward. OPD now appears across frontier post-training stacks~\citep{yang2025qwen3,zeng2026glm,xiao2026mimo,xu2026deepseek,blakeman2026nemotron}, several of which adopt the multi-teacher form we study in Section~\ref{sec:mopd}, routing each query to a domain specialist. A growing line explains why OPD is effective, and does so entirely on the algorithm side. One thread studies its objective and update geometry: the dense per-token target replaces the single sequence-level scalar of outcome-reward RL, which in parameter space steers an early-locked, low-rank update subspace~\citep{cai2026learning,shen2026geometry} and can be recast as token-level policy gradient~\citep{ko2026scaling}, connecting to analyses that separate the dense updates of SFT from the localized subnetworks of RL~\citep{mukherjee2026reinforcement,zhu2025path}. A second thread charts when OPD succeeds or fails: a thinking pattern shared between teacher and student and the limits of an over-strong teacher~\citep{li2026rethinking}, which tokens are actually learnable~\citep{armandpour2026unmasking,wang2026not}, and recurring failure modes with simple fixes~\citep{zhu2026many,fu2026revisiting}. These accounts take the training set as given. We instead ask how many training inputs OPD needs, and find that the binding resource is not the data but the rate at which the student absorbs it.

\paragraph{Data-efficient reasoning post-training.}
A parallel line studies how little data elicits strong reasoning. Under sparse outcome-reward RL, \citet{wang2025oneshot} show that a single example keeps improving a model for over a thousand steps, the precedent we adapt to the dense-signal setting of OPD. LIMR~\citep{limr} prunes the RLVR training set several-fold by ranking each example's learning impact. With supervised fine-tuning, LIMA~\citep{zhou2023lima}, LIMO~\citep{ye2025limo}, and s1~\citep{muennighoff2025s1} elicit competitive alignment or reasoning from on the order of a thousand curated examples. Within OPD, scheduling supervision over reasoning prefixes cuts training cost severalfold~\citep{zhang2026fast}. Broader post-training work selects high-value subsets by influence or quality~\citep{xia2024less,chen2024alpagasus,he2025right,he2025air}. These works share our data-efficiency theme, but their supervision still depends on curated problems, delivered as a sparse outcome scalar in the RL case or as reference solutions in the SFT case. OPD instead supplies a dense per-token target at each visited prefix and remains effective on queries the student never solves (Section~\ref{sec:one-shot}), where the teacher signal stays informative even on incorrect rollouts~\citep{armandpour2026unmasking}, so the limiting factor moves from whether a useful signal is available, the standing difficulty under sparse reward, to how fast an on-policy student absorbs an already-abundant one (Section~\ref{sec:mech-dynamics}).

\paragraph{Synthetic and unsupervised post-training data.}
A growing body of work reduces post-training's dependence on curated, human-written data. One direction synthesizes the inputs: Self-Instruct~\citep{wang2023self} and Evol-Instruct~\citep{xu2024wizardlm} bootstrap instructions from a language model, and Magpie~\citep{xu2025magpie} elicits full instruction--response pairs from nothing but a chat-template prefix. A second direction removes ground-truth labels from RL~\citep{he2026far}, replacing verifiable rewards with self-generated signals such as model confidence or entropy~\citep{zhao2026learning,agarwal2026unreasonable} and majority agreement across rollouts~\citep{zuo2026ttrl}, or drawing reward from unlabeled corpora~\citep{dong2025reinforcement}. A third removes external problems altogether, letting the model propose its own tasks~\citep{zhao2026absolute,zweiger2026self}. These lines relax supervision on the reward or the query source. Our template- and WildChat-OPD results (Section~\ref{sec:template}) push the query side further: because the teacher supplies dense supervision at each visited prefix, the query's role reduces to placing the student in a useful reasoning state, so an off-domain chat log or even a content-free template drives OPD nearly as well as a curated problem.
\section{Conclusion, Limitations, and Future Work}
\label{sec:conclusion}
\label{sec:limitations}

\paragraph{Conclusion.} 
Motivated by one-shot RLVR~\citep{wang2025oneshot}, we asked whether a single query is also enough for OPD, and it is: one-shot OPD keeps the student improving for hundreds of steps and recovers most of full-data OPD's gain, across model families and task domains. The data and the algorithm sides give one answer: OPD is data-overfed but algorithm-starved. A query acts on the student through the states its rollouts reach, and one query already covers most of the state space full-data OPD visits, while 16 semantically diverse queries match full-data training in single-domain OPD and in MOPD alike. What limits a run is the absorption rate: the proportion of the remaining teacher--student gap one update absorbs keeps falling, at the same pace on one query as on all 17k, so the optimizer rather than the training set sets how long a run takes. An input need only start the student reasoning, which turns data design from collecting problems into choosing teachers.

\paragraph{Limitations.}
State coverage is a semantic-level proxy measured against a reference space built from full-data rollouts, so it reports how much of that space a query set reaches rather than what it covers on its own, and it weights every cluster equally, however often the cluster is visited and however much teacher signal it still carries. What sets the absorption rate is still open. Our MOPD runs use three domains with one teacher each, leaving open how far the diversity result extends as teachers are added.

\paragraph{Future work.}
Three directions follow.
\begin{itemize}[topsep=2pt, partopsep=0pt, leftmargin=12pt, itemsep=2pt]
    \item Selecting data by state coverage: the question is no longer how many queries to collect but which states they induce, so state coverage becomes a selection criterion once it can be estimated from the queries themselves, without a full-data run to measure it against.
    \item Making training more step-efficient, by reusing a batch for several epochs under a trust region on the per-token gap, or weighting tokens by how much teacher signal they still carry.
    \item Growing the setting, to MOPD with more teachers and domains, and to larger students, agentic tool use, and long contexts, where training data is most expensive to collect.
\end{itemize}

\bibliography{main}

\newpage
\clearpage
\appendix
\section{Details for Section~\ref{sec:one-shot}}

\subsection{Experimental Setup and Training Queries}
\label{app:model-data}

\paragraph{Instruction-following setup.}
\label{app:IF-setup}
Unlike the other primary-domain teachers, the instruction-following teacher is trained specifically for our experiments. Starting from DeepSeek-R1-Distill-Qwen-1.5B, we post-train the model on synthetic instruction-following tasks with programmatically checkable constraints. We use GRPO with rule-based outcome rewards that indicate whether the response satisfies the target-turn constraints. Training samples four completions per prompt at temperature \(0.9\), uses a maximum prompt length of \(4{,}096\) tokens and a maximum response length of \(16{,}384\) tokens, and optimizes with AdamW at learning rate \(1\times10^{-6}\), train batch size \(16\), and clip ratio \(0.2\). The step-600 checkpoint is used as the instruction-following teacher, improving IFBench \citep{pyatkin2026generalizing} from \(14.33\) to \(20.65\) and Multi-IF final-turn accuracy from \(20.84\) to \(28.58\).

For this domain, ``full-data'' denotes the subset sampled from UltraData-SFT-2605 \citep{ultradata-sft-2605} rather than the complete dataset, and the one-shot query is drawn at random from that subset.

\begin{figure}[t]
    \centering
    \includegraphics[width=0.45\linewidth]{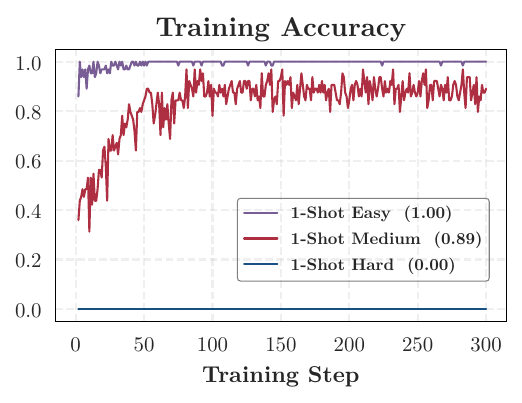}
    \caption{Fraction of the 64 rollouts per step that solve the training query itself, for the three queries of Table~\ref{tab:one-shot-queries}.}
    \label{fig:probe_prompt_training_acc}
\end{figure}

\paragraph{Mathematical queries for one-shot OPD (Table~\ref{tab:one-shot-queries}, Figure~\ref{fig:probe_prompt_training_acc}).}
\label{app:training-accuracy}
We sample 8 rollouts from the untrained student for each of the first 64 queries of DAPO-Math-17K and assign difficulty by the resulting pass rate, which selects \texttt{t59} (\(8/8\)), \texttt{t22} (\(4/8\)), and \texttt{t56} (\(0/8\)) as the easy, medium, and hard queries. How often each is solved during training follows that assignment: the easy query stays near \(1.0\) throughout, the medium one rises from \(0.36\) to about \(0.89\), and the hard one is never solved in \(300\) steps, yet all three produce comparable benchmark gains in Figure~\ref{fig:probe_prompt}.

\begin{table}[t]
    \centering
    \small
    \setlength{\tabcolsep}{4pt}
    \begin{tabular}{p{0.10\textwidth}p{0.08\textwidth}p{0.74\textwidth}}
    \toprule
    Difficulty & ID & Query \\
    \midrule
    Easy & \texttt{t59} &
    In triangle \(ABC\), let \(D\) be the foot of the altitude from \(A\). Suppose that \(AD=4\), \(BD=3\), \(CD=2\), and \(AB\) is extended past \(B\) to a point \(E\) such that \(BE=5\). Determine the value of \(CE^2\). \\
    \midrule
    Medium & \texttt{t22} &
    Let \(S\) be the set of triples \((a,b,c)\) of non-negative integers such that \(a+b+c\) is even. Determine the value of
    \(\sum_{(a,b,c)\in S} \frac{1}{2^a3^b5^c}\).
    This sum can be expressed as \(\frac{m}{n}\), where \(m\) and \(n\) are relatively prime positive integers. Compute \(m+n\). \\
    \midrule
    Hard & \texttt{t56} &
    Byan is playing a game called ``raven, raven, falcon'' with his three friends. His friends sit in a circle, and Byan walks clockwise around them, tapping each friend he passes on the head and saying either ``raven'' or ``falcon,'' each with probability \(\frac{1}{2}\). The game ends when Byan has said ``falcon'' twice. The probability that one of his friends will be called a ``falcon'' twice can be expressed as \(\frac{m}{n}\), where \(m\) and \(n\) are relatively prime positive integers. Find \(100m+n\). \\
    \bottomrule
    \end{tabular}
    \caption{Mathematical queries used in the query-difficulty experiment, labelled by the student's pass rate before OPD training.}
    \label{tab:one-shot-queries}
\end{table}

\paragraph{Default hyperparameters for OPD (Table~\ref{tab:opd-hyperparams}).}
Unless otherwise noted, all experiments use these defaults. The top-\(K\) row covers both objectives: the mathematical one-shot runs of Section~\ref{sec:finding-recovery} use \(K=16\), and all other experiments use the sampled-token objective.

\begin{table}[t]
\centering
\caption{Default hyperparameters for OPD.}
\label{tab:opd-hyperparams}
\small
\begin{tabular}{ll}
\toprule
\textbf{Item} & \textbf{Value} \\
\midrule
Rollout batch size      & 64 \\
Mini batch size         & 64 \\
Responses per prompt    & 1 \\
KL coefficient          & 0.0 \\
LogProb top-$K$         & 0(default) / 16 \\
Top-$K$ strategy        & Student Top-$K$ \\
Loss aggregation        & token-mean \\
Training temperature    & 1.0 \\
Top-$p$                 & 1.0 \\
Optimizer               & AdamW ($\beta_1{=}0.9$, $\beta_2{=}0.999$, wd $=0.01$) \\
Learning rate           & 1e-6 \\
Gradient clip norm      & 1.0 \\
Max prompt length       & 1024 (math); 4096 (code, IF, agentic) \\
Max response length     & 7168 (math, code, IF); 2048 (agentic) \\
\bottomrule
\end{tabular}
\end{table}

\section{Details for Section~\ref{sec:data}}

\subsection{Training-State Coverage}
\label{app:state-coverage}
\label{app:state-coverage-robustness}

\paragraph{Construction.}
Rollouts are collected by optimization step rather than at random, so that every stage of training is equally represented and the budget is comparable across settings, which run for the same number of steps: at each step from \(1\) to \(500\) of the full-data run we take \(5\) rollouts, three building the clusters and two forming full data (held-out), for \(12{,}000\) reference and \(8{,}000\) held-out states. PCA reduces these to \(50\) dimensions, retaining \(71.9\%\) of the variance, before \(K\)-means. Every evaluated setting is then measured on that same schedule, drawing \(2\) of the \(5\) rollouts at each step and accumulating over the run, so the step-\(300\) budget of \(600\) rollouts matches the held-out split. We repeat the draw \(30\) times; bands and error bars are one standard deviation over these repetitions, which reflects rollout subsampling rather than variation across trained models.

\begin{figure}[t]
    \centering
    \includegraphics[width=0.85\linewidth]{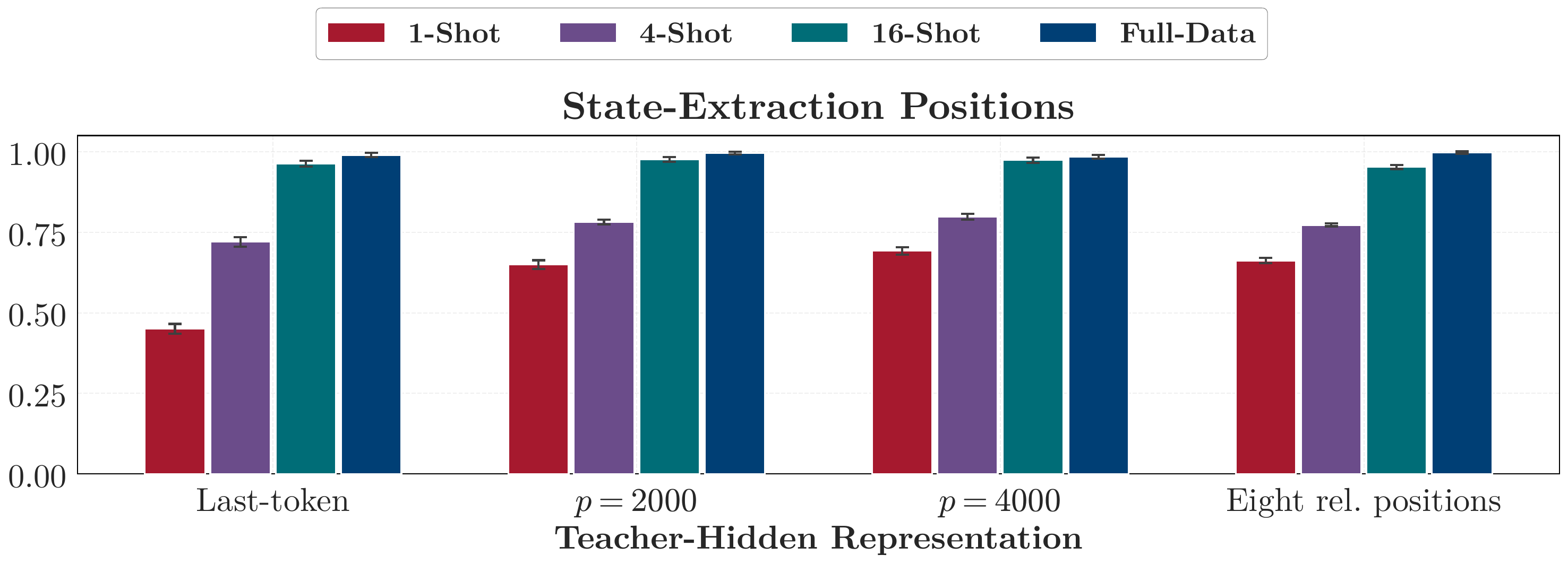}
    \caption{Training-state coverage under alternative state-extraction positions, at \(K=200\). Error bars show one standard deviation over rollout subsamples.}
    \label{fig:state-position-sensitivity}
\end{figure}

\paragraph{The ordering is stable across \(K\), reference construction, and state positions (Tables~\ref{tab:coverage-k} and~\ref{tab:coverage-reference}, Figure~\ref{fig:state-position-sensitivity}).}
Coverage depends on how finely the reference states are clustered and on which states define the clusters, so we vary both. The first table repeats the analysis at \(K\in\{50,100,200,500\}\): finer clusters lower absolute coverage, but coverage still increases from 1- to 4- to 16-shot with 16-shot close to full data. The second rebuilds the clusters from the union of \(2{,}000\) rollouts per setting instead of from full-data states alone, and the ordering survives; because states from all settings help define these pooled clusters, pooled coverage is not a recovery measure, and its rows are to be read only against each other. The figure varies the positions read from each rollout, comparing the final non-padding token and fixed absolute positions \(2{,}000\) and \(4{,}000\) against the 8 relative positions of the main analysis. All four preserve the ordering. The fixed-position variants land close to the main construction, whereas the final token gives markedly lower coverage when there are few queries, which is what one expects of a position governed by response length and termination; we therefore read 8 relative positions, which sample throughout a response of any length.

\begin{table}[t]
    \centering
    \small
    \caption{Full-data-anchored training-state coverage at step \(300\) under different cluster counts \(K\).}
    \label{tab:coverage-k}
    \begin{tabular}{lcccc}
        \toprule
        \(K\)
        & 1-shot
        & 4-shot
        & 16-shot
        & Full data (held-out) \\
        \midrule
        50  & 0.755 & 0.829 & 1.000 & 1.000 \\
        100 & 0.727 & 0.803 & 1.000 & 1.000 \\
        200 & 0.715 & 0.798 & 0.989 & 1.000 \\
        500 & 0.564 & 0.695 & 0.904 & 0.956 \\
        \bottomrule
    \end{tabular}
\end{table}

\begin{table}[t]
    \centering
    \small
    \caption{Training-state coverage under alternative cluster constructions, at \(K=200\).}
    \label{tab:coverage-reference}
    \begin{tabular}{lcccc}
        \toprule
        Reference construction
        & 1-shot
        & 4-shot
        & 16-shot
        & Full data (held-out) \\
        \midrule
        Full-data anchored
        & 0.663 & 0.773 & 0.953 & 0.998 \\
        Equal-size pooled
        & 0.750 & 0.845 & 0.958 & 0.966 \\
        \bottomrule
    \end{tabular}
\end{table}

\paragraph{One-shot has a heavier tail of distant states, but is not separable from full data (Table~\ref{tab:reference-diagnostics}).}
Nearest-centroid assignment places a state in some cluster however far it lies from every full-data state, so coverage alone cannot tell a setting that fills the reference space thinly from one that leaves it. We therefore also measure, for each state, the mean cosine distance to its ten nearest reference states in the unreduced \(1536\)-dimensional space, and call a state off-reference when that distance exceeds the \(95\)th percentile of the held-out full-data states. One-shot OPD exceeds the threshold more often than the other settings, so its distances have a heavier upper tail. But an AUROC of \(0.552\) against held-out full-data states is close to chance, so one-shot states are not separable from full-data states as a whole.

\begin{table}[t]
    \centering
    \small
    \caption{Reference-distance diagnostics across query-set sizes. AUROC compares each setting with the held-out full-data states. The last row is the complete full-data pool, including the rollouts that built the clusters, so its exceedance rate need not equal the \(5\%\) the threshold is set at.}
    \label{tab:reference-diagnostics}
    \begin{tabular}{lccc}
        \toprule
        Setting
        & Exceedance (\%)
        & Mean distance
        & AUROC \\
        \midrule
        1-shot   & 17.2 & 0.283 & 0.552 \\
        4-shot   & 4.4  & 0.218 & 0.477 \\
        16-shot  & 3.8  & 0.231 & 0.512 \\
        Full data (all)  & 4.0  & 0.215 & 0.476 \\
        \bottomrule
    \end{tabular}
\end{table}

\subsection{Controls on the Training Data}
\label{app:fixed-count-diversity}
\label{app:schedule}

\begin{figure}[t]
  \centering
  \includegraphics[width=0.75\linewidth]{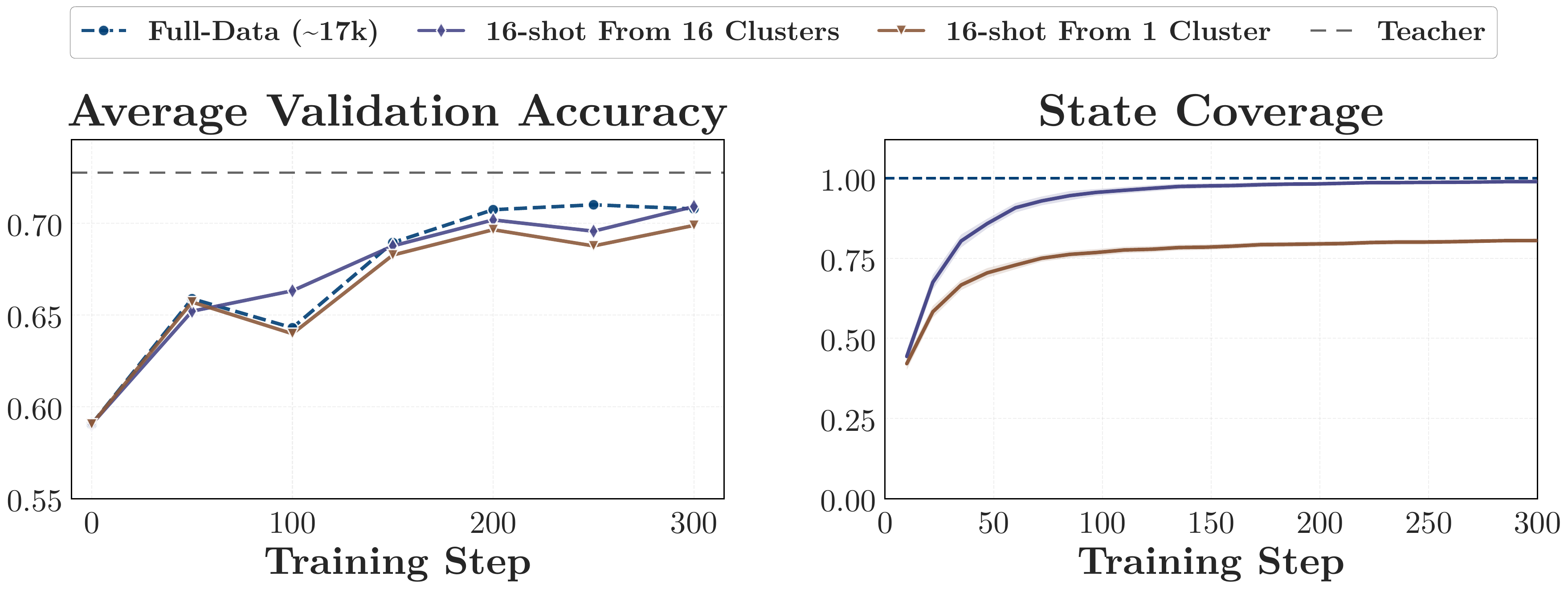}
  \caption{Two 16-shot query sets that differ only in diversity, against full-data OPD.}
  \label{fig:fixed-count-diversity}
\end{figure}

\paragraph{Diversity in 16-shot OPD (Figure~\ref{fig:fixed-count-diversity}).}
Every query on the ladder of Section~\ref{sec:state-diversity} represents a different semantic cluster, so it raises diversity and count together. Holding the count at 16 and drawing the second set from a single cluster separates them: the 16 cluster representatives reach \(70.9\) validation accuracy at step \(300\) against \(69.9\) for the single-cluster set, which is about what 4-shot OPD reaches with a quarter as many queries, and their state coverage separates the same way and does so early, \(98.9\%\) against a single-cluster set that flattens near \(76.8\%\) within \(100\) steps. Queries from one cluster keep returning the student to the part of the state space it has already seen.

\begin{figure}[t]
  \centering
  \includegraphics[width=0.75\linewidth]{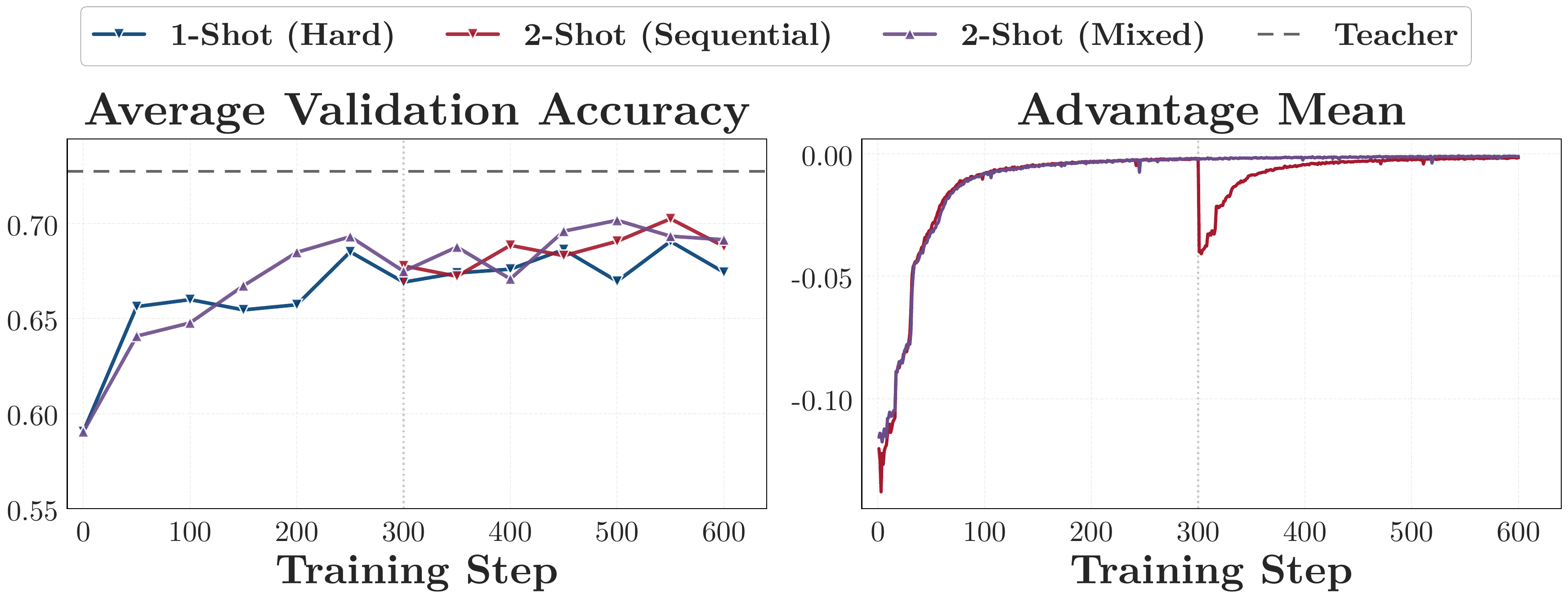}
  \caption{Effect of query scheduling in 2-shot OPD. The vertical dotted line marks the query switch in the sequential schedule.}
  \label{fig:nshot_schedule}
\end{figure}

\paragraph{Query scheduling in 2-shot OPD (Figure~\ref{fig:nshot_schedule}).}
Holding a two-query set fixed and varying only when each query is shown, the sequential schedule trains on the hard query for the first \(300\) steps of a \(600\)-step budget and on the medium query for the rest, while the mixed schedule puts \(32\) copies of each in every batch. At step \(600\) they reach \(68.8\) and \(69.1\), both above the \(67.4\) of the one-shot baseline they share, so ordering a fixed query set barely matters next to what the second query itself buys. These runs train longer than those of Section~\ref{sec:state-diversity} and take the hard query as their baseline, so neither their horizon nor their baseline is interchangeable with Figure~\ref{fig:state-diversity}.

\section{Details for Section~\ref{sec:algorithm}}

\subsection{Estimating the Absorption Rate and the Measurement Window}
\label{app:alignment-dynamics}

Section~\ref{sec:mech-dynamics} measures under three conventions. Why each one:
\begin{itemize}[topsep=2pt, partopsep=0pt, leftmargin=13pt, itemsep=2pt, parsep=0pt]
    \item \(\boldsymbol{[t/1.4,\,1.4t]}\). A difference between adjacent steps is mostly noise once \(d\) moves slowly, so \(v_t\) is read as the slope of \(\log d\) across the endpoints of this window, using the median \(d\) near each. Only centers whose window fits inside the measured range are plotted, which is why the main text reads the rate at steps \(50\) and \(200\) rather than at step \(300\).
    \item \(\boldsymbol{d_t/d_{30}}\). Each run measures its distance on the states it visits and the four sit at different levels by step \(30\), so plotting each against its own value there lets the curves be compared by shape rather than by level.
    \item \textbf{Start at step \(30\).} Gradient clipping binds until then, and a clipped update is not the update the objective asked for: \(v_t\) inside that window measures what one \emph{fixed-size} step buys rather than what one step buys, which is why the measured rate rises there rather than falls.
\end{itemize}

Table~\ref{tab:gap-reduction} collects the readings Section~\ref{sec:mech-dynamics} quotes from this range: the proportion of the step-\(30\) distance still left at step \(300\), and the factor by which the absorption rate falls between steps \(50\) and \(200\).

\begin{table}[t]
    \centering
    \small
    \caption{Alignment dynamics over steps \(30\)--\(300\) for on-policy OPD, at the default learning rate.}
    \label{tab:gap-reduction}
    \begin{tabular}{lcccc}
        \toprule
        Setting (queries) & 1 & 4 & 16 & Full data \\
        \midrule
        \(d_{300}/d_{30}\) & 0.16 & 0.17 & 0.18 & 0.22 \\
        \(v_{50}/v_{200}\) & 5.6 & 5.2 & 4.2 & 6.4 \\
        \bottomrule
    \end{tabular}
\end{table}

\subsection{The Learning Rate Rescales the Step Axis}
\label{app:lr-sensitivity}

We repeat on-policy one-shot OPD at half, the default, and twice the default learning rate, holding every other setting fixed, and read all three over the same \(300\) steps as Section~\ref{sec:mech-dynamics}.

\begin{figure}[t]
    \centering
    \includegraphics[width=\linewidth]{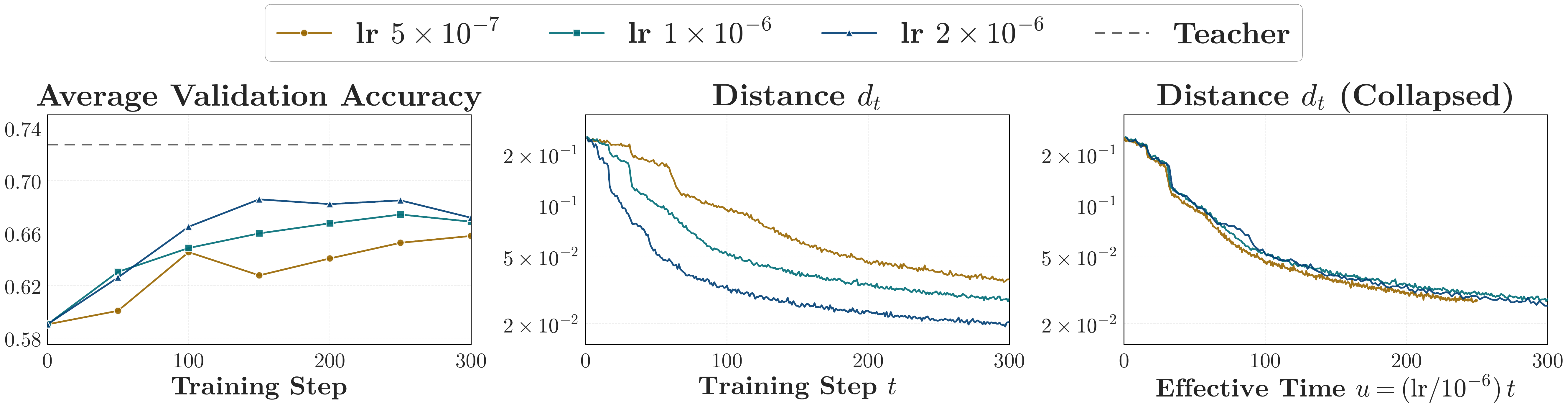}
    \caption{Validation accuracy and teacher--student distance for one-shot OPD at half, one, and twice the default learning rate. Distance is on a logarithmic vertical axis against linear time.}
    \label{fig:lr-sensitivity}
\end{figure}

\paragraph{The learning rate rescales the step axis and leaves the slowdown untouched.}
A larger learning rate reaches any given distance sooner, the three curves separated by roughly the ratio of the learning rates (Figure~\ref{fig:lr-sensitivity}). Changing the horizontal axis from the step \(t\) to \(u=(\mathrm{lr}/10^{-6})\,t\) brings them onto one curve: the spread between the three runs narrows from about \(2.2\times\) to about \(1.1\times\). The learning rate therefore sets only how fast a run moves along that one curve, not the shape of the curve itself. It bends flatter as it goes, just as the four query settings of Section~\ref{sec:mech-dynamics} do against the step count. A proportionally larger learning rate therefore shortens the run proportionally, over the fourfold range we tested, without reaching whatever makes alignment slow in the first place.

\section{Details for Section~\ref{sec:mopd}}
Table~\ref{tab:mopd-hyperparams} shows the default hyperparameters for MOPD.

\begin{table}[!ht]
\centering
\caption{Default hyperparameters for MOPD (multi-teacher on-policy distillation).}
\label{tab:mopd-hyperparams}
\small
\begin{tabular}{ll}
\toprule
\textbf{Item} & \textbf{Value} \\
\midrule
Number of teachers      & 3 (math / code / IF) \\
LogProb top-$k$         & 0  \\
KL coefficient          & 0.0 \\
Loss aggregation        & seq-mean-token-mean \\
\midrule
Rollout batch size      & 64 \\
Mini batch size         & 64 \\
Responses per prompt    & 1 \\
Training temperature    & 1.0 \\
Top-$p$                 & 1.0 \\
\midrule
Optimizer               & AdamW ($\beta_1{=}0.9$, $\beta_2{=}0.999$, wd $=0.01$) \\
Learning rate           & 1e-6 \\
Gradient clipping       & 1.0 \\
\midrule
Max prompt length       & 4096 \\
Max response length     & 7680 \\
\bottomrule
\end{tabular}
\end{table}

\section{Details for Section~\ref{sec:discussion}}

\subsection{Template and WildChat OPD in Code Generation}
\label{app:template-code}

Section~\ref{sec:template} reports the four input conditions in mathematics. Here we report the same four in code generation.

\paragraph{Setup.}
The student is R1-Distill-1.5B, the teacher DeepCoder-1.5B-Preview~\citep{deepcoder2025}, and the baseline full-data OPD on TACO~\citep{li2023taco}. The base template, the system template, and the WildChat queries are exactly those of Section~\ref{sec:template}; the system template's one-line instruction is the competitive-programming instruction of Figure~\ref{fig:template-examples}b. We evaluate on LiveCodeBench v6 under avg@3.

\begin{figure}[!htb]
  \centering
  \includegraphics[width=\textwidth]{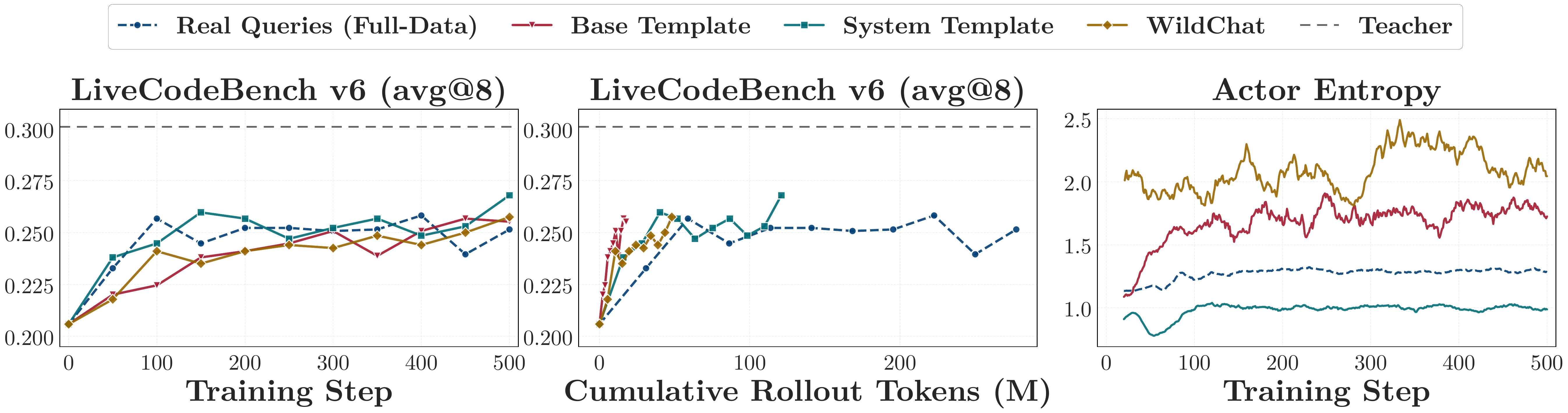}
  \caption{Validation accuracy and actor entropy for code-generation OPD.
  % trained on the real training set, on a content-light template, or on general-domain WildChat queries.
  }
  \label{fig:template-wildchat-code}
\end{figure}

\paragraph{Results.}
Figure~\ref{fig:template-wildchat-code} shows the four conditions ending within \(1.7\) points of one another, and the token accounting separating them much further than the step accounting does: over the same \(500\) steps the baseline spends \(277\)M rollout tokens against \(18\)M for the base template, with the other two in between. Actor entropy orders the conditions exactly as it does in mathematics, with WildChat highest and the system template lowest. The one difference is that no condition falls behind here, whereas WildChat costs about a point in mathematics.

\end{document}